\documentclass[lettersize,journal]{IEEEtran}
\usepackage{array}
\usepackage{textcomp}
\usepackage{stfloats}
\usepackage{verbatim} 
\usepackage{balance}
\usepackage{amsthm,amsmath,amsfonts,amssymb,bm} 
\usepackage{epsfig}
\usepackage[parfill]{parskip} 
\usepackage{bookmark}
\usepackage{url}   
\usepackage{mathtools}

\usepackage{physics}
\usepackage{textcomp} 
\usepackage{csquotes} 
\usepackage[dvipsnames]{xcolor} 
\usepackage{graphicx} 
\usepackage{epstopdf} 
\graphicspath{{./plot/}}

\usepackage{hyperref} 
\usepackage{cleveref} 
\usepackage{caption} 

\usepackage{siunitx}
\ifCLASSOPTIONcompsoc
  \usepackage[nocompress]{cite}
\else
  \usepackage{cite}
\fi

\ifCLASSOPTIONcompsoc
\usepackage[caption=false,font=footnotesize,labelfont=sf,textfont=sf]{subfig}
\else
 \usepackage[caption=false,font=footnotesize]{subfig}
\fi

\makeatletter
\def\url@leostyle{%
  \@ifundefined{selectfont}{\def\UrlFont{\sf}}{\def\UrlFont{\small\ttfamily}}}
\makeatother
\usepackage{mathrsfs}
\usepackage{tabularx} 
\newcolumntype{b}{X}
\newcolumntype{s}{>{\hsize=.8\hsize}X}
\usepackage{multirow} 
\usepackage{float} 
\usepackage{cleveref} 
\usepackage{subfiles} 
\usepackage{hyperref} 
\usepackage{adjustbox}
\usepackage{lineno}

\usepackage{algorithm}
\usepackage[noend]{algpseudocode}
\makeatletter
\newcommand{\multiline}[1]{%
  \begin{tabularx}{\dimexpr\linewidth-\ALG@thistlm}[t]{@{}X@{}}
    #1
  \end{tabularx}
}
\makeatother

\usepackage{multicol}
\usepackage{nccmath}
\theoremstyle{plain}
\newtheorem*{theorem*}{Theorem}
\newtheorem{theorem}{Theorem}

\usepackage{url}

\begin{document}
\title{Information Theoretical Importance Sampling Clustering}

\author{Jiangshe~Zhang,
        Lizhen~Ji,
        and~Meng~Wang
\IEEEcompsocitemizethanks{
\IEEEcompsocthanksitem
Jiangshe Zhang and Lizhen Ji is with School of Mathematics and Statistics, 
Xi'an Jiaotong University, Xi'an, Shaanxi, China, 710049.\protect \hfil\break
E-mail: jszhang@mail.xjtu.edu.cn, jlz\_stat@stu.xjtu.edu.cn. \hfil\break
Meng Wang is with Northwest China Grid Company Limited.\protect \hfil\break
E-mail: wangmeng@nw.sgcc.com.cn \hfil\break
Corresponding author:Jiangshe Zhang \hfil\break}}
\maketitle

\begin{abstract}
A current assumption of most clustering methods is that the training data and future data are taken from the same distribution.
However, this assumption may not hold in most real-world scenarios.
In this paper, we propose an information theoretical importance sampling based approach for clustering problems (ITISC)
which minimizes the worst case of expected distortions under the constraint of distribution deviation.
The distribution deviation constraint can be converted to the constraint
over a set of weight distributions centered on the uniform distribution derived from importance sampling.
The objective of the proposed approach is to minimize the loss under maximum degradation hence
the resulting problem is a constrained minimax optimization problem
which can be reformulated to an unconstrained problem using the Lagrange method.
The optimization problem can be solved by both an alternative optimization algorithm
or a general optimization routine by commercially available software.
Experiment results on synthetic datasets and a real-world load forecasting problem 
validate the effectiveness of the proposed model.
Furthermore, we show that fuzzy c-means is a special case of ITISC with the logarithmic distortion,
and this observation provides an interesting physical interpretation for fuzzy exponent $m$.
\end{abstract}


\section{Introduction}\label{introduction}
The problems of clustering aim at the optimal grouping of the observed data and appear in very diverse fields 
including pattern recognition, data mining and signal compression\cite{xu2005survey}.
K-means\cite{duda1973pattern,krishna1999genetic} is one of the most simple and popular clustering algorithms 
in many machine learning research\cite{nie2022effective}.
However, it tends to get trapped in a local minimum\cite{rose1990deterministic}.
Fuzzy c-means (FCM) \cite{dunn1973fuzzy, bezdek1973fuzzy, bezdek2013pattern, bezdek1984fcm} 
is one of the most widely used objective function based clustering methods
which assigns each data point to multiple clusters with some degree of sharing.
Sadaaki et al.\cite{sadaaki1997fuzzy} show that the update rule for fuzzy membership of FCM
is the same as maximum entropy approach under logarithmic transformation of distortion,
but the calculation of the centers are different between the two methods.
Deterministic annealing\cite{rose1990deterministic,rose1993constrained}
technique is also proposed for the nonconvex optimization problem of clustering 
in order to avoid local minima of the given cost function.
Both FCM and deterministic annealing clustering start with an attempt to alleviate the local minimum trap problem 
suffered by hard c-means \cite{duda1973pattern} and achieved better performance in most cases.
The deterministic annealing clustering has a clear physical interpretation,
but we haven't yet found a solid theoretical foundation for FCM, 
and the parameter $m$ involved seems unnatural without any physical meaning. 

Another crucial assumption underlying most current theory of clustering problem is that the distribution
of training samples is identical to the distribution of future test samples, but it is often violated in practice
where the distribution of future data deviates from the distribution of training data.
For example, a decision-making system forecasts future actions in the presence of 
current uncertainty and imperfect knowledge\cite{garibaldi2019need}.
In this paper, we propose a clustering model based on importance sampling
which minimizes the worst case\cite{farnia2016minimax} of expected distortions under the constraint of distribution deviation.
The distribution deviation is measured by the Kullback-Leibler 
divergence\cite{kullback1951information,williams1980bayesian,sadaaki1997fuzzy,ichihashi2000gaussian,
coppi2006fuzzy,ortega2013thermodynamics,genewein2015bounded,hihn2019information}
between the current distribution and a future distribution.
The proposed model is called the \textit{Information Theoretical Importance Sampling Clustering}, 
denoted as ITISC for short.
We show that ITISC is an extension of FCM,
and provide a physical interpretation for the fuzzy component $m$.  

The proposed ITISC method aims to minimize the loss in maximum degradation
and hence the resulting optimal problem is a minimax problem.
Inspired from the importance sampling method\cite{tokdar2010importance,shi2009neural,shi2009hierarchical}, 
we convert the constraint between the current distribution and a future distribution
to a constraint on the importance sampling weights. 
The constrained minimax problem can be reformulated to an unconstrained problem using the Lagrange method
with two parameters (Lagrange multipliers).
One controls the fuzziness of membership of clusterings, and we call it the ``fuzziness temperature".
The other controls the deviation of distribution shifts, and we call it the ``deviation temperature". 
The advantage of the reformulation of ITISC is that the 
resulting unconstrained optimization problem is dependent on cluster center only
and the solution to the corresponding optimization problem can be found by 
applying the quasi-newton algorithms\cite{gill1972quasi,fletcher2013practical}
or alternative optimization algorithm.

We conduct experiments on both Gaussian synthetic datasets and
a real-world load forecasting dataset to validate the effectiveness of ITISC. 
First, an evaluation metric called M-BoundaryDist is proposed 
as a measure of how well a clustering algorithm performs
with respect to the boundary points.
M-BoundaryDist calculates the sum of distances of boundary points to the dataset centroid.
Then, we explain why the boundary points matter in clustering problems.
Traditional robust clustering methods\cite{dave1997robust,zhang2003robust} claim the 
points far away from the data as noisy points or outliers,
and a robust clustering algorithm ought to find the cluster centers even in the presence of outliers. 
However, in some cases, all observable data points are important with no outliers. 
For example, when establishing a postal system, we need to consider not only the operating revenue
but also the fairness among different regions, 
even if the operating revenue in urban areas is higher compared with rural areas.
In this case, we can not regard remote area as ``outliers''. 
Our proposed ITISC method assumes that all observable data points are important with no outliers.
Experiment results on Gaussian synthetic datasets show that
when the deviation temperature is small, the cluster centers of ITISC are closer to the boundary points
compared with k-means, FCM and hierarchical clustering\cite{ward1963hierarchical}
and performs better under large distribution shifts.
Next, results on a load forecasting problem show that 
ITISC performs better compared with k-means, FCM and hierarchical clustering
on 10 out of 12 months on future load series. 
Both synthetic and real-world examples validate the effectiveness of ITISC.

\textbf{Outline of the paper.} 
\Cref{Related Work} gives a brief review on FCM and importance sampling. 
\Cref{ITISC} describes our proposed information theoretical importance sampling clustering model
and the algorithm to solve it. 
Specifically, \Cref{ITISC-model} formalizes the ITISC model. 
\Cref{ITISC-algorithm} reformulates the constrained optimization problem of ITISC into 
an unconstrained optimization problem. 
\Cref{algorithm} proposes two algorithms to solve the optimization problem.
\Cref{Fuzzy-ITISC} presents Fuzzy-ITISC, which uses the logarithmic transformation of the distortion.
A physical interpretation for the fuzzy component $m$ is also revealed
in this section. 
\Cref{Results} presents numerical results on synthetic data set 
to demonstrate the effectiveness of the proposed Fuzzy-ITISC.
\Cref{load-forecasting} applies Fuzzy-ITISC on a real-world load forecasting problem and 
show that Fuzzy-ITISC outperforms other compared clustering algorithms under most scenarios of future distribution shifts.  
Finally, we conclude this paper in \Cref{conclusion}.

\section{Related Work} \label{Related Work}
Let $\mathcal{D} =\{x_1,x_2,\cdots,x_N\}$ be a given set of $N$ points in $S$ dimensional space.
These data points are to be partitioned into $C$ clusters.
We denote the prototype of cluster $j$ as $y_j$.
$Y=\{y_1, y_2, \cdots, y_C\}$ denotes all cluster centers and 
$d(x_i, y_j)$ denotes the \textbf{squared} distance between $x_i$ and $y_j$,
which is usually used as the distortion measure.

\textbf{Importance Sampling}
Importance sampling\cite{glynn1989importance,tokdar2010importance,robert2010introducing}
refers to a collection of Monte Carlo methods
where a mathematical expectation with respect to an unknown distribution
is approximated by a weighted average of random draws from another known distribution.
For a discrete random variable $X$ with probability mass function $p(x)$,
i.e., $p(x)=P(X=x)$. We aim to compute $\mu = E(f(X))$.
Suppose $q(x)$ is another discrete distribution 
such that $q(x)=0$ implies $f(x)p(x)=0$, 
then the expectation $\mu_f$ is computed as follows 
\begin{equation}
  \mu_f = E_{p}(f(X)) = \sum_x f(x) p(x)  
        = \frac{\sum_x \frac{p(x)f(x)}{q(x)}q(x)}{\sum_x \frac{p(x)}{q(x)} q(x) }.
\end{equation}
Suppose $\{x_1, x_2, \dots, x_N \}$ are $N$ i.i.d. samples 
drawn from $q(x)$, then the empirical estimator for $\mu_f$ is
\begin{equation}
  \hat{\mu}_f = \frac{\frac{1}{N} \sum_{i=1}^{N} \frac{p(x_i)}{q(x_i)}f(x_i)}{\frac{1}{N} \sum_{l=1}^{N} \frac{p(x_l)}{q(x_l)}}
   = \sum_{i=1}^{N} w(x_i) f(x_i),
\end{equation}
where 
\begin{equation} 
w(x_i) = \frac{\frac{p(x_i)}{q(x_i)}}{\sum_{l=1}^{N} \frac{p(x_l)}{q(x_l)}},\label{eq:IS-weight}
\end{equation}
\eqref{eq:IS-weight} is called the importance sampling weight in this paper.

\textbf{Fuzzy c-means}
Fuzzy clustering is a fruitful extension of hard c-means\cite{duda1973pattern}
with various applications and is supported by cognitive evidence. 
The fuzzy clustering algorithms regard each cluster as a fuzzy set
and each data point may be assigned to multiple clusters 
with some degree of sharing\cite{menard2004non}.
In FCM\cite{bezdek1984fcm}, an exponent parameter $m$ is introduced and 
$u_{ij} $ is interpreted as the fuzzy membership with values in $[0,1]$
which measures the degree to which the $i$-th data point belongs to the $j$-th cluster. 
The corresponding objective function $F_{FCM}(Y,U)$ and the constraints are as follows
\begin{align}
  \begin{split} 
  \label{eqn:FCM}
  \min_{Y,U}   \quad &   F_{FCM}(Y,U)=\sum_{i=1}^{N}\sum_{j=1}^{C} u_{ij}^{m} d(x_i,y_j)\\
  \text{s. t.} \quad &  \sum_{j=1}^{C} u_{ij}=1, 1\leq i \leq N \\
                     & 0 < \sum_{i=1}^{N} u_{ij} < N, 1\leq j \leq C \\
                     & u_{ij}\in [0,1], 1\leq i \leq N, 1\leq j \leq C,
  \end{split}
\end{align}
where $m \in [1, \infty)$ is a fuzzy exponent called the fuzzifier.
The larger $m$ is, the fuzzier the partition\cite{ichihashi2000gaussian}.
$F_{FCM}(Y,U)$ is alternatively optimized by 
optimizing $U$ for a fixed cluster parameters and 
optimizing $Y$ for a fixed membership degrees.
The update formulas are obtained by setting the derivative $F_{FCM}(Y,U)$ with respect to 
parameters $U$, $Y$ to zero\cite{bezdek2013pattern}, which is
\begin{align}
u_{ij} & =\frac{d(x_i,y_j)^{\frac{1}{1-m}}}{\sum_{j=1}^{c}d(x_i,y_j)^{\frac{1}{1-m}}}, 
       \quad 1\leq j \leq C, \quad 1\leq i \leq N. \label{eq:FCM-Uij} \\
y_j    & = \frac{\sum_{i=1}^{N} u_{ij}^{m} y_j}{\sum_{i=1}^{N} u_{ij}^{m}},
       \quad \quad \quad \quad 1 \leq j \leq C. \label{eq:FCM-Y} 
\end{align}
Another way to solve \eqref{eqn:FCM} is by the reformulated criteria of FCM\cite{hathaway1995optimization}.
Substituting  \eqref{eq:FCM-Uij} into \eqref{eqn:FCM}, we get
\begin{align}
R_{FCM}(Y) = \sum_{i=1}^{N} (\sum_{j=1}^{C} d(x_i,y_j)^{1\over 1-m})^{1-m}. \label{eq:FCM-reform}
\end{align}
The function $F_{FCM}(Y,U)$ depends on both $U$ and $Y$ and the function $R_{FCM}(Y)$ depends on $Y$ only.
The aim of reformulation is to decrease the number of variables by eliminating $U$
by the optimal necessary condition with respect to $U$.
\eqref{eq:FCM-reform} can be solved by commercially available software.
The underlying assumption of traditional clustering methods 
is that the distribution of training data is the same as future data,
however it may not hold in most real cases. 
In the following section, we propose a new clustering algorithm 
to handle this problem derived from the importance sampling method.

\section{Information theoretical importance sampling clustering} \label{ITISC}
In the proposed information theoretical importance sampling clustering method,
we assume that the observed data set draws from a distribution $q(x)$
and our aim is to construct a clustering algorithm for a population with unknown distribution $p(x)$.
We further assume that if $p(x)$ are, instead of being completely unknown,
restricted to a class of distributions, i.e.
\begin{equation}
  \Gamma = \{p(x): KL(p(x)||q(x)) \leq C_1 \}. \label{eq:gamma}
\end{equation}
A \textit{minimax} approach is applied through minimizing the 
worst-case loss restricted to this constraint.
\Cref{ITISC-model} gives a principled derivation of the minimax approach and
\Cref{ITISC-algorithm} solves the corresponding optimization problem.
The derivation of our proposed approach in this paper is heavily dependent on the work\cite{rose1998deterministic}.

\subsection{Principle of ITISC} \label{ITISC-model}
In this section, we present the principled deviation of the minimax approach.
In our proposed algorithm, we aim to minimize the \textit{worst-case situation}
of expected distortion under the given constraints. 
Let $x$ denote a data point or source vector, $y(x)$ denote its representation cluster center, 
and $d(x,y(x))$ denote the distortion measure. 
For a random variable $X$ with distribution $p(x)$, the expected distortion for this representation can be written as
\begin{equation}
D = \sum_x\sum_y p(x,y)d(x,y), \label{eq:Loss}
\end{equation} 
where $p(x,y)$ is the joint probability distribution
and $p(y|x)$ is the association probability relating input vector $x$ and cluster center $y$. 
Here $d(x,y)$ represents a \textbf{squared} distance between $x$ and $y$ for convenience.
The following derivation also holds for any other distortion measures. 

First, following \cite{rose1990deterministic,rose1998deterministic},
we find the best partition $U$ to minimize the expected distortion $D$
subject to a specified level of randomness.
The level of randomness is usually measured by the joint entropy $H(X,Y)$,
which can be decomposed into sums of entropy and conditional entropy, $H(X,Y)=H(X) + H(Y|X)$.
Since $H(X)$ is independent of clustering, we use the conditional entropy $H(Y|X)$ instead.
Then, the corresponding optimization problem becomes 
\begin{align}
  \begin{split}
  \label{eqn:DA}
  \min_{p(y|x)} \quad & D = \sum_x\sum_y p(x,y)d(x,y) \\ 
  \text{s.t.}   \quad & H(Y|X) \leq C_0.
  \end{split}
\end{align}
\eqref{eqn:DA} is the optimization problem for deterministic annealing 
clustering\cite{rose1990deterministic,rose1998deterministic}.
An equivalent derivation can be obtained by the principle of maximum entropy\cite{jaynes1957information}
in which the level of expected distortion is fixed\cite{rose1998deterministic}.
Second, for a given partition $p(y|x)$ , we find the $p(x)$
which maximizing the objective function which corresponds to the \textit{worst-case situation}.
However, $p(x)$ is unknown in the problem and we assume that
$p(x)$ is subject to the constraint \eqref{eq:gamma}.
Therefore, the corresponding optimization problem becomes
\begin{align}
  \begin{split}
  \max_{p(x)} \min_{p(y|x)} \quad & D = \sum_x\sum_y p(x,y)d(x,y)  \\ 
  \text{s.t.} \quad & H(Y|X) \leq C_0 \\
              \quad & KL(p(x)||q(x)) \leq C_1.
  \end{split}
\end{align}
Third, given the fuzzy partition $p(y|x)$ and the worst-case distribution $p(x)$,
we aim to find the best prototype $y$ which minimizes the objective function. 
Then the corresponding optimization problem is   
\begin{align}
  \begin{split} \label{eq:L-with-constraints}
  \min_{y} \max_{p(x)} \min_{p(y|x)} \quad &  D = \sum_x\sum_y p(x,y)d(x,y) \\
  \text{s.t.} \quad & H(Y|X) \leq C_0 \\
  \quad & KL(p(x)||q(x)) \leq C_1.
  \end{split}
\end{align}
\eqref{eq:L-with-constraints} is the formalized optimization problem of ITISC.
Next, we get the empirical estimation of the expected distortion, 
the conditional entropy and the KL divergence based on the importance sampling technique.
Suppose $Y=\{y_1,y_2,\cdots,y_C\}$ is a finite set and 
the observed dataset $\mathcal{D} =\{x_1,x_2,\cdots,x_N\}$ are $N$ i.i.d. samples drawn from $q(x)$.
The importance sampling weight for $x_i$ is denoted as $w_i$,
and $W=[w_i]_{N \times 1}$ with $\sum_{i=1}^{N} w_i=1$.
The association probability $p(y_j|x_i)$ is denoted as $u_{ij}$ 
and the fuzzy membership matrix is $U=[u_{ij}]_{N \times C}$.
Then the empirical estimate of $D$ is
\begin{equation}
  \hat{D} = \sum_{i=1}^{N} w_i \sum_{j=1}^{C} u_{ij} d(x_i, y_j), \label{eq:L-empirical}
\end{equation}
the empirical estimate of $H(Y|X)$ is
\begin{equation}
  \hat{H}(Y|X) =  \sum_{i=1}^{N} w_i \sum_{j=1}^{C} u_{ij} log u_{ij}, \label{eq:H(Y|X)-empirical}
\end{equation}
and the empirical estimate of $KL(p(x)\parallel q(x))$ is 
\begin{equation} 
    \widehat{KL}(p(x) \parallel q(x)) = KL(w(x_i) \parallel \{\frac{1}{N}\})  
    = \sum_{i=1}^{N} w_i log w_i + log N. \label{eq:KL-pq-empirical}
\end{equation}
The derivations of \eqref{eq:L-empirical}, \eqref{eq:H(Y|X)-empirical} and \eqref{eq:KL-pq-empirical}
are shown in Appendix-B.
Following this, the constrained optimization problem \eqref{eq:L-with-constraints}  can be reformulated to 
the unconstrained optimization problem using the Lagrange method
\begin{equation}
  F_{ITISC}^{0}(Y,W,U) = D-T_1H(Y|X) -T_2 KL(w(x_i)||\{  \frac{1}{N} \}), \label{eq:ITISC-objective} \\
\end{equation}
where $T_1 > 0$ and $T_2 > 0$ are the temperature parameters.
$T_1$ is the fuzziness temperature which governs the level of randomness of the conditional entropy.
When $T_2=0$, the problem degenerates to the deterministic annealing for clustering\cite{rose1990deterministic},
which is put forward for the nonconvex optimization problem of clustering
in order to avoid local minima of the given cost function.
$T_2$ is the deviation temperature which governs the level of randomness of the distribution deviation.
When $T_2 \rightarrow 0$, the distribution shift $KL(p(x)) \parallel q(x))$ can be very large 
and for $T_2 \rightarrow \infty $, the distribution shift should be small,
the effect of $T_2$ is further illustrated in \Cref{T2}.
Plugging \eqref{eq:L-empirical}, \eqref{eq:H(Y|X)-empirical} and \eqref{eq:KL-pq-empirical} 
back into \eqref{eq:ITISC-objective}, we get the empirical estimates of the objective function for ITISC
clustering, which is
\begin{align}
  F_{ITISC}(Y,W,U) & = \sum_{i=1}^{N} w_i\{\sum_{j=1}^{C} u_{ij} d(x_i,y_j)  \label{eq:ITISC-empirical} \\
  & + T_1\sum_{j=1}^{C}u_{ij}log u_{ij}\} - T_2\sum_{i=1}^{N} w_i log w_i. \nonumber
\end{align}
Here we omitting the last term $-T_2log(N)$ for simplification
since $T_2$ is predefined and the last term $log(N)$ is a constant. 
Adding back the constraints on the partition matrix $U$ and the importance sampling weight $W$,
we finally get the optimization problem of ITISC, which is
\begin{align}
  \min_{Y} \max_{W} \min_{U} \quad & F_{ITISC}(Y,W,U) \label{eq:ITISC} \\
  \text{s.t.} \quad  & \sum_{j=1}^{C} u_{ij}=1, 1 \leq i \leq N \nonumber \\
  & 0 < \sum_{i=1}^{N} u_{ij} < N, 1 \leq j \leq C \label{eq:constraints-uij} \\
  & u_{ij} \in [0,1], 1\leq i \leq N ,1 \leq j \leq C  \nonumber \\
  & \sum_{i=1}^{N}w_{i}=1, w_i \in [0,1], 1 \leq i \leq N,  \label{eq:constraints-wi}
\end{align}
where \eqref{eq:constraints-uij} is the constraints for the fuzzy membership $U$
and \eqref{eq:constraints-wi} is the constraints for the importance sampling weight $W$.
In conclusion, ITISC is an objective-function-based clustering method and 
the objective function can be seen as a trade-off between 
the expected distortion, the level of randomness and the distribution deviation. 

\subsection{Reformulation of ITISC} \label{ITISC-algorithm}
In this section, we give a reformulation of the optimization problem of the ITISC model. 
We derive the membership and the weight update equations
from the necessary optimality conditions for minimization of the cost function
by differentiating $F_{ITISC}(U,W,Y)$ with respect to $U$, $W$ and set the derivatives to zero. 
Specifically, let the Lagrange multipliers be $\{\lambda_i \}_{i=1}^{N}$ and $\lambda$,
then the Lagrange function becomes
\begin{align}
\mathcal{L}_{ITISC} & = \sum_{i=1}^{N} w_i\{\sum_{j=1}^{C} u_{ij} d(x_i,y_j)+
                        T_1\sum_{j=1}^{C}u_{ij}log u_{ij}\} \nonumber \\
                   & -T_2\sum_{i=1}^{N} w_i log w_i \nonumber \\ 
                   & - \sum_{i=1}^{N} \lambda_i (\sum_{j=1}^{C} u_{ij} -1) - \lambda (\sum_{i=1}^{N} w_i-1). \label{eq:ITISC-lagrange}
\end{align}
Setting the derivative $\mathcal{L}_{ITISC}$ with respect to $U$ to zero, 
we get the optimality necessary condition for $U$, which is 
\begin{equation}    
  u_{ij} =  \frac{exp(-\frac{d(x_i, y_j)}{T_1})}{\sum_{k=1}^{C} exp(-\frac{d(x_i, y_k)}{T_1})}. \label{eq:ITISC-uij}
\end{equation}
Plugging \eqref{eq:ITISC-uij} back into \eqref{eq:ITISC-lagrange}, we get the reformulation for $U$, which is  
\begin{align}
& R_{ITISC}(Y,W) = -T_1 \sum_{i=1}^{N} w_i [log \sum_{j=1}^{C} exp(-\frac{d(x_i, y_j)}{T_1})] \nonumber \\
& \quad -T_2\sum_{i=1}^{N} w_i log w_i - \lambda (\sum_{i=1}^{N} w_i-1). \label{eq:ITISC-Y-W}
\end{align}
Setting the derivative $R_{ITISC}(Y,W)$ with respect to $W$ to zero, 
we get the optimality necessary condition for $W$, which is 
\begin{equation}
  w_i = \frac{[\sum_{j=1}^{C}  exp(-\frac{d(x_i, y_j)}{T_1})]^{-\frac{T_1}{T_2}}}
  {\sum_{l=1}^{N} [\sum_{j=1}^{C}  exp(-\frac{d(x_l, y_j)}{T_1})]^{-\frac{T_1}{T_2}}}.\label{eq:ITISC-wi}
\end{equation}
Substituting  \eqref{eq:ITISC-wi} into \eqref{eq:ITISC-Y-W}, we get the reformulation for $U$ and $W$, which is 
\begin{equation}
R_{ITISC}(Y) = T_2 log(\sum_{l=1}^{N} [\sum_{j=1}^{C} exp(-\frac{d(x_i, y_j)}{T_1})]^{-\frac{T_1}{T_2}}). \label{eq:ITISC-Y}
\end{equation}
We call $R_{ITISC}(Y)$ the reformulation function of $F_{ITISC}(Y,W,U)$
and the minimization of $R_{ITISC}(Y)$ with respect to $Y$ is equivalent to 
the min-max-min of $F_{ITISC}(Y,W,U)$ with respect to $Y,W,U$. 
Therefore, finding the solution to ITISC becomes minimization of 
$R_{ITISC}(Y)$ with respect to $Y$.
$R_{ITISC}(Y)$ can be seen as the effective cost function when representing $x_i$ with the cluster prototype $y_j$. 
The derivations of \Crefrange{eq:ITISC-uij}{eq:ITISC-Y} are shown in Appendix-C. 

\textit{Remark:}
ITISC model can be seen as a two-level statistical physical model.
For the first system, for a given $x_i$, 
if we regard $d(x_i,y_j)$ as the energy for the prototype $y_j$, then
\begin{equation}
  \sum_{j=1}^{C} u_{ij} d(x_i,y_j) + T_1\sum_{j=1}^{C}u_{ij}log u_{ij} \label{eq:HFE}
\end{equation}
becomes the Helmholtz free energy with the temperature $T_1$\cite{rose1998deterministic}.
In bounded rationality theory\cite{genewein2015bounded},  
\begin{equation}
-T_1 log \sum_{j=1}^{C} exp(-\frac{d(x_i,y_j)}{T_1}) \label{eq:CE}
\end{equation}
\eqref{eq:CE} is called the certainty equivalence.
For the second system, if we regard
\begin{equation}
log [\sum_{j=1}^{C} exp(-\frac{d(x_i, y_j)}{T_1})]^{T_1} \label{eq:energyX}
\end{equation}
as the energy for $x_i$, then
\begin{equation}
- \sum_{i=1}^{N} w_i [log \sum_{j=1}^{C} exp(-\frac{d(x_i, y_j)}{T_1})]^{T_1} -T_2 \sum_{i=1}^{N} w_i log w_i
\end{equation}
becomes the negative Helmholtz free energy with the temperature $T_2$.
In the following section, we present two algorithms 
to solve the optimization problem in \eqref{eq:ITISC-Y}. 

\subsection{Algorithms} \label{algorithm}
In this section, we present two algorithms to solve the optimization problem of the ITISC model.
The first algorithm follows the \textbf{A}lternative \textbf{O}ptimization algorithm for FCM\cite{bezdek2013pattern}, 
therefore denoted as ITISC-AO.
The second algorithm follows the \textbf{R}eformulation algorithm for FCM\cite{hathaway1995optimization},
therefore denoted as ITISC-R.
\subsubsection{ITISC-AO}
The first method is optimizing $R_{ITISC}(Y)$ via simple Picard iteration 
by repeatedly updating $U$, $W$ and $Y$ until a given stopping criterion is satisfied.
Specifically, setting the derivative $R_{ITISC}(Y)$ with respect to $y_k$ to zero, we get the optimality condition for $Y$
\begin{equation}
  \sum_{i=1}^{N} w_i u_{ik} \frac{\partial}{\partial y_k}d(x_i,y_k)=0.\label{eq:ITISC-yj-necc}
\end{equation}
For Euclidean distance, the update rule for the center $y_k$ is 
\begin{equation}
  y_k = \frac{\sum_{i=1}^{N} w_i u_{ik} x_i}{\sum_{i=1}^{N} w_i u_{ik}}. \label{eq:ITISC-yj}
\end{equation}
The derivations of \eqref{eq:ITISC-yj-necc} and \eqref{eq:ITISC-yj} are shown in Appendix-C. 
We formalize the Information Theoretical Importance Sampling Clustering Alternative Optimization(ITISC-AO) 
algorithm in \autoref{alg:ITISC-AO}.
\begin{algorithm}[H] 
\caption{ITISC-AO} \label{alg:ITISC-AO}
\begin{algorithmic}[1]
  \renewcommand{\algorithmicrequire}{\textbf{Input:}}
  \renewcommand{\algorithmicensure}{\textbf{Output:}}
  \Require X,C,$T_1$,$T_2$
  \Ensure  U,W,Y 
  \State Sample initial $U$ and $W$ from the standard uniform distribution
         and normalize them column-wise to satisfy the normalization constraint 
         \eqref{eq:constraints-uij} and \eqref{eq:constraints-wi}.
  \State Choose $C$ centers uniformly at random from $\mathcal{X}$.
  \While{$\norm{Y^{t+1}-Y^{t}} > \epsilon$}
  \State Apply \eqref{eq:ITISC-uij} to compute $U$.
  \State Apply \eqref{eq:ITISC-wi}  to compute $W$.
  \State Apply \eqref{eq:ITISC-yj-necc} to compute $Y$. 
         For squared Euclidean distance, use \eqref{eq:ITISC-yj} to update $y_j$.
  \EndWhile 
\end{algorithmic}
\end{algorithm}
The superscript $t$ in Step-3 represents the $t$-th iteration.
One existing problem for ITISC-AO is that when $T_2$ is small, the algorithm may not converge. 
This question remains an open question under current investigation and will be studied in the future. 
\newline

\subsubsection{ITISC-R}
The second method is optimizing $R_{ITISC}(Y)$ directly, which is similar to \cite{hathaway1995optimization}
which uses \textit{fminunc's} BFGS\cite{fletcher2013practical,gill1972quasi} algorithm
in MATLAB Optimization Toolbox\cite{MatlabOTB}. 
We formalize the Information Theoretical Importance Sampling Clustering Reformulation(ITISC-R) algorithm 
in \autoref{alg:ITISC-R}.
\begin{algorithm}[H]
  \caption{ITISC-R} \label{alg:ITISC-R}
  \begin{algorithmic}[1]
    \renewcommand{\algorithmicrequire}{\textbf{Input:}}
    \renewcommand{\algorithmicensure}{\textbf{Output:}}
    \Require $X$,$C$,$T_1$,$T_2$
    \Ensure  $U$,$W$,$Y$ 
    \State Sample initial $U$ and $W$ from the standard uniform distribution,
           normalize $U$ and $W$ column-wise to satisfy the normalization constraint 
           \eqref{eq:constraints-uij} and \eqref{eq:constraints-wi}.
    \State Choose $C$ centers uniformly at random from $\mathcal{X}$.
    \State Use \textit{fminunc} in MATLAB optimization Toolbox to minimize \eqref{eq:ITISC-Y} w.r.t. $Y$ 
           until a given stopping criteria is satisfied. 
    \State Apply \eqref{eq:ITISC-uij} to compute $U$.
    \State Apply \eqref{eq:ITISC-wi} to compute $W$.
\end{algorithmic}
\end{algorithm}
In Step-3 of \autoref{alg:ITISC-R}, we use the default optimality tolerance 
$\Delta F_{ITISC}(Y) \leq 10^{-6}$ in MATLAB as the stopping criteria.

\subsection{Fuzzy-ITISC} \label{Fuzzy-ITISC}
In this section, we use the \textbf{logarithmic} transformation\cite{sadaaki1997fuzzy} of
$d(x,y)$ as the distortion measure and denote the resulting ITISC method as Fuzzy-ITISC.
Similar to ITISC, the optimization problem for Fuzzy-ITISC is 
\begin{align}
  \begin{split} \label{eq:L-log-with-constraints}
  \min_{y} \max_{p(x)} \min_{p(y|x)} \quad &  D^{fuzzy} = \sum_x\sum_y p(x,y)logd(x,y) \\
  \text{s.t.} \quad & H(Y|X) \leq C_0 \\
  \quad & KL(p(x)||q(x)) \leq C_1.
  \end{split}
\end{align}
The optimality necessary conditions for $U$, $W$, $Y$ and the reformulation for Fuzzy-ITISC are as follows
\begin{align}
  u_{ij} &  = \frac{d(x_i, y_j)^{-\frac{1}{T_1}}}{\sum_{k=1}^{C} d(x_i, y_k)^{-\frac{1}{T_1}}}, \label{eq:Fuzzy-ITISC-uij} \\
  w_i    & = \frac{[\sum_{j=1}^{C} d(x_i, y_j)^{-\frac{1}{T_1}}]^{-\frac{T_1}{T_2}}}
                {\sum_{l=1}^{N} [\sum_{j=1}^{C} d(x_l, y_j)^{-\frac{1}{T_1}}]^{-\frac{T_1}{T_2}}}, \label{eq:Fuzzy-ITISC-wi} \\
  \sum_{i=1}^{N} & (w_i)^{(1-T_2)} (u_{ik})^{(1+T_1)}\frac{\partial}{\partial y_k}d(x_i,y_k) = 0, \label{eq:Fuzzy-ITISC-yj-necc} \\
  R^{fuzzy}_{ITISC}(Y) &= T_2 log  (\sum_{i=1}^{N} (\sum_{j=1}^{C} d(x_i, y_j)^{-\frac{1}{T_1}})^{-\frac{T_1}{T_2}}). \label{eq:Fuzzy-ITISC-Y}
\end{align}
The iterative update rule for the cluster center $y_k$ under the Euclidean distance is 
\begin{equation}
  y_k = \frac{\sum_{i=1}^{N} (w_i)^{1-T_2} (u_{ik})^{1+T_1} x_i}
                {\sum_{i=1}^{N} (w_i)^{1-T_2} (u_{ik})^{1+T_1}}. \label{eq:Fuzzy-ITISC-yj}
\end{equation}
The derivations of \eqref{eq:Fuzzy-ITISC-yj-necc} and \eqref{eq:Fuzzy-ITISC-yj} are shown in Appendix-D.
The algorithm using \eqref{eq:Fuzzy-ITISC-uij}, \eqref{eq:Fuzzy-ITISC-wi} and \eqref{eq:Fuzzy-ITISC-yj-necc}
to update $U$, $W$ and $Y$ respectively is denoted as Fuzzy-ITISC-AO. 
The algorithm which directly minimizes the reformulation 
$R_{ITISC}(Y)$ with respect to $Y$ using MATLAB is denoted as Fuzzy-ITISC-R.
%
Finally, we present the important finding in this paper which reveals the  
physical interpretation for fuzzy exponent $m$.
Let $T_1=m-1$, $T_2=1$, then \eqref{eq:Fuzzy-ITISC-Y} becomes
\begin{align}
  R^{fuzzy}_{ITISC}(Y)= log (\sum_{i=1}^{N} (\sum_{j=1}^{C} d(x_i, y_j)^{\frac{1}{1-m}})^{1-m}).\label{eq:ITISC-FCM}
\end{align}
Comparing \eqref{eq:ITISC-FCM} with the reformulation of FCM
\begin{equation}
R_F(Y)=\sum_{i=1}^{N}(\sum_{j=1}^{C}d(x_i,y_j)^{1\over 1-m})^{1-m}, \label{eq:FCM-reformulation}
\end{equation}
we can see that the minimization of $R_{F}(Y)$ with respect to $Y$ is equivalent to 
the minimization of $R_{ITISC}^{fuzzy}(Y)$ with respect to $Y$. 
In conclusion, we obtain the following theorem which 
reveals the relationship between FCM and Fuzzy-ITISC.
\begin{theorem} \label{Fuzzy-ITISC-FCM}
The FCM is a special case of ITISC in which distortion is measured by $log d(x,y)$  
and the parameters $T_1$, $T_2$ are set as $T_1=m-1$, $T_2=1$.
\end{theorem}
Therefore, \textbf{the fuzzy component} $\bm{m=T_1+1}$ \textbf{in FCM can be interpreted  
as the recalibration of temperature in thermodynamic system}.
\autoref{Fuzzy-ITISC-FCM} reveals there is a deep relationship between FCM
and thermodynamics\cite{rose1998deterministic}.

\section{Numerical Results on Synthetic data set} \label{Results}
In this section, we present numerical results on synthetic data set 
to demonstrate the effectiveness of the proposed Fuzzy-ITISC.
Specifically, we present the clustering results of Fuzzy-ITISC in \Cref{synthetic},
which reveal that the centers of the Fuzzy-ITISC are closer to the boundary points. 
In \Cref{ISWeights}, we analyze the importance sampling weights.  
In \Cref{T2}, we investigate how the temperature $T_2$ affects the Fuzzy-ITISC results.
Finally, in \Cref{dist-shift}, we show that Fuzzy-ITISC outperforms k-means, FCM and
hierarchical clustering under large distribution shifts.

\textbf{Experiment settings}
In this section, we use the Gaussian synthetic datasets with 2,3,4 and 6 clusters,
with each cluster contains 200 points that are normally distributed over a two-dimensional space. 
The details of these datasets can be found in Appendix-A. 
The \textit{default} dataset is a Gaussian synthetic dataset with 3 clusters.
We follow the python package scikit\cite{scikit-learn} for the implementation of k-means
and use k-means++\cite{arthur2006k} as the initialization method.
As for  hierarchical clustering, we use the ward linkage\cite{ward1963hierarchical}
if not otherwise specified.
In later discussions, the Hierarchical Clustering is abbreviated as HC for ease of reference.
We use \cite{dias2019fuzzy} for the implementation of FCM and use 
the commonly chosen $m=2$  as the default value in all compared FCM models.
We set $T_1=1.0$ as the default value for all Fuzzy-ITISC models
since the temperature $T_1$ is analyzed in detail in deterministic annealing clustering\cite{rose1998deterministic}
and it behaves similarly in Fuzzy-ITISC.
In this paper, we conduct all experiments on Fuzzy-ITISC model
using the logarithmic distortion $logd(x,y)$ and the squared Euclidean distance.
For the optimization method, we use the Fuzzy-ITISC-R 
as the default optimization method if not otherwise specified.

\subsection{Clustering Results of Fuzzy-ITISC} \label{synthetic}
In this section, we first propose a metric called M-BoundaryDist
as a measure of how well a clustering algorithm performs
with respect to the boundary points.
Then, we compare the clustering results of Fuzzy-ITISC, k-means, FCM and HC on synthetic datasets.

\textbf{M-BoundaryDist}
Suppose the \textit{centroid} of one dataset is the mean of the data across each dimension. 
The \textit{boundary points} are the points far away from the centroid of the dataset. 
We denote the centroid of the dataset as $D_{centroid}$ and 
the $M$ boundary points as \textit{M-BoundaryPoints}.
Suppose the boundary points assigned to the cluster-$j$ are denoted as $x^{j}_1, \ldots ,x^{j}_{c_j}$,
where $c_j$ is the number of boundary points assigned to the cluster-$j$
and $y_j$ represents the cluster center.
Next, M-BoundaryDist is defined as follows
\begin{equation}
  \text{M-BoundaryDist} = \sum_{j=1}^{C} \sum_{m=1}^{c_j} d(x^{m}_j,y_j). \label{eq:M-BoundaryDist}
\end{equation}
Clearly, $\sum_{j=1}^{C}c_{j}=M$.
When $M=1$, the boundary point is called MaxBoundaryPoint and the corresponding
distance is called MaxBoundaryDist.
\begin{figure}
  \begin{center}
  \includegraphics[width=.9\linewidth]{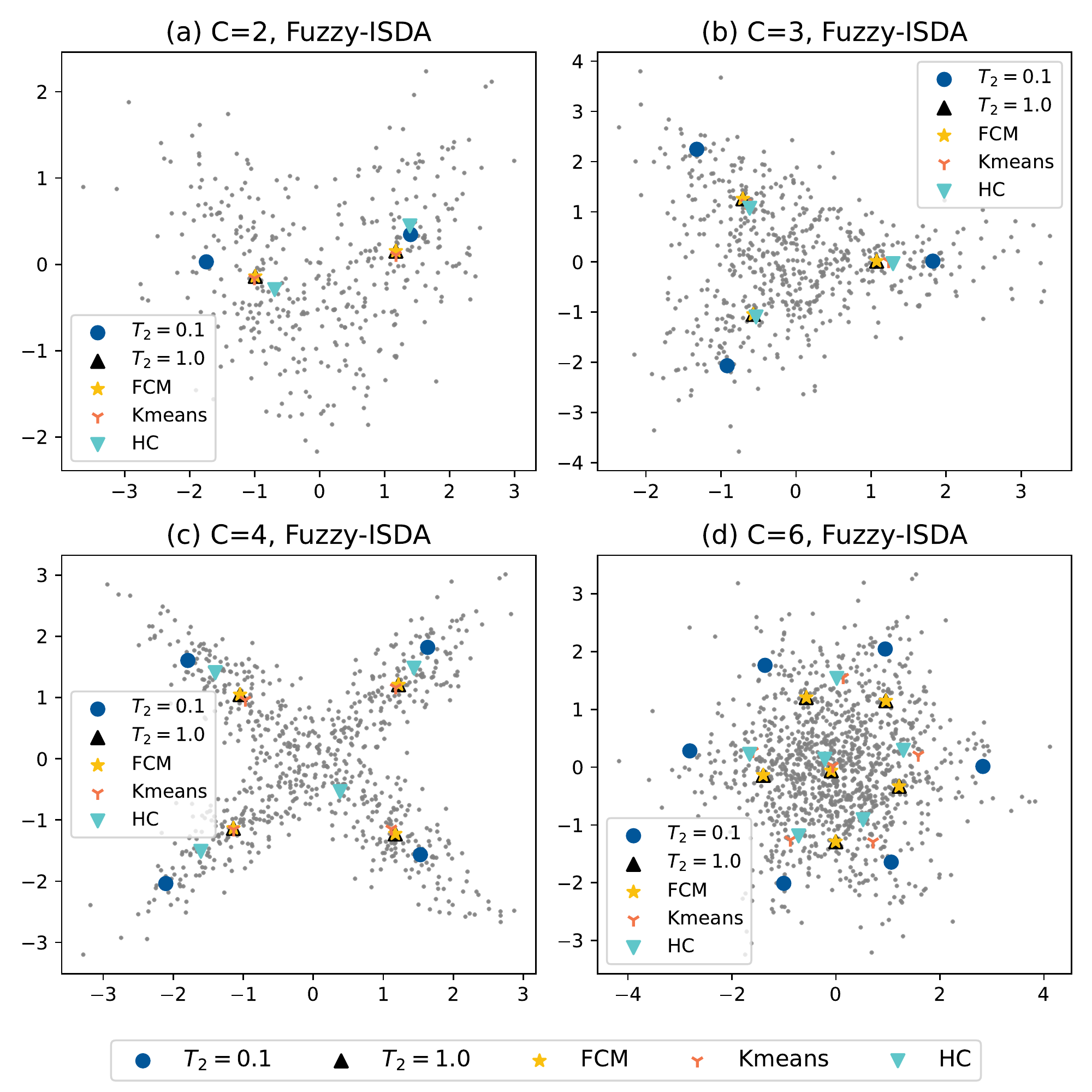}
  \end{center}
\caption{Clustering results comparison between k-means, FCM, HC, Fuzzy-ITISC($T_2=1.0$) and Fuzzy-ITISC($T_2=0.1$)
on four Gaussian synthetic datasets with 2,3,4,6 clusters.}
\label{fig:Fuzzy-ITISC-4-examples}
\end{figure}
\begin{table}
  \caption{\label{tab:MBD}
  Comparison of Max-BoundaryDist among different models and different datasets.
  $C$ means the number of clusters and Extreme means the extreme dataset. 
  KM represents k-means and FI represents Fuzzy-ITISC.}
  \centering
  \small
  \begin{tabular}{lccccc}
    \hline
      Data & KM & FCM & HC & FI($T_2=0.1$) & FI($T_2=1.0$)\\
    \hline
      C=2&2.84&2.84&3.17&2.08&2.84\\
      C=3&2.98&2.88&3.09&1.72&2.88\\
      C=4&2.96&2.98&2.38&1.66&2.98\\
      C=6&2.58&2.78&2.52&1.37&2.78\\
      Extreme&4.41&7.28&4.41&2.27&7.29\\    
    \hline
  \end{tabular}
\end{table}

The clustering result for four synthetic datasets with 2,3,4,6 clusters 
are shown in \autoref{fig:Fuzzy-ITISC-4-examples}(a),(b),(c),(d) respectively.
The figure shows that the centers of Fuzzy-ITISC are closer
to the boundary points compared to the other clustering methods.
Addtionally, the figure shows that the centers of 
Fuzzy-ITISC($T_2=1.0$) overlap with those of FCM($m=2.0$),
which validates the result in \autoref{Fuzzy-ITISC-FCM}.
\autoref{tab:MBD} compares the Max-BoundaryDist among the five models and four datasets.
The table indicates that the Max-BoundaryDist of Fuzzy-ITISC($T_2=0.1$)
is smaller than that of k-means, FCM and HC. 
This observation suggests that Fuzzy-ITISC performs better than the other models
in terms of Max-BoundaryDist.
In other words, Fuzzy-ITISC \textit{takes care of} boundary points when $T_2$ is small.

\begin{figure}
  \begin{center}
  \includegraphics[width=.9999\linewidth]{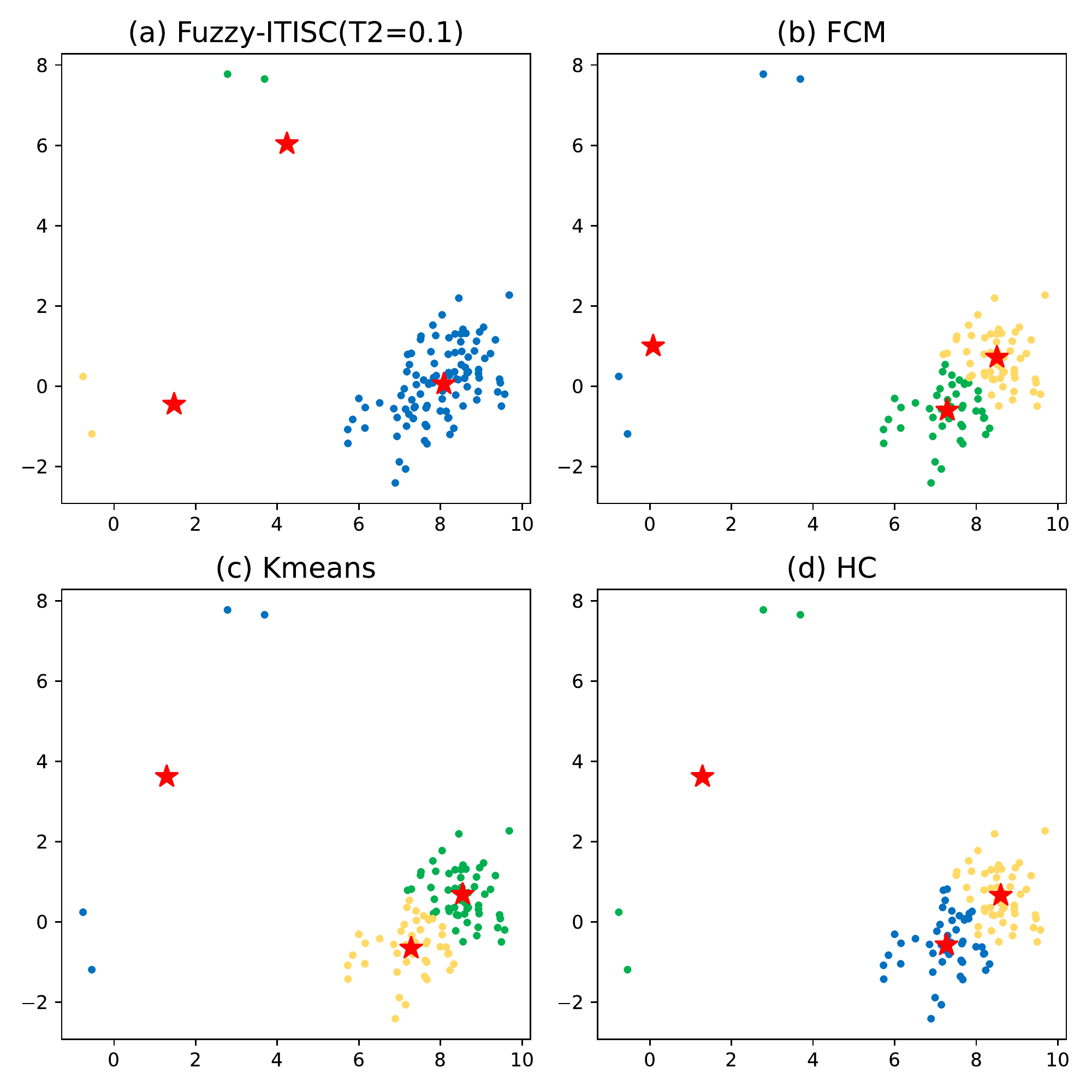}
  \end{center}
\caption{Comparison of Fuzzy-ITISC, FCM, k-means and HC of an extreme case. 
Stars represent cluster centers.}
\label{fig:extreme}
\end{figure}

To demonstrate the differences between Fuzzy-ITISC and the other clustering methods,
we examine an \textit{Extreme} dataset with three clusters.
The Gaussian synthetic dataset has cluster means of [1,0], [8,0] and [4,8]
and covariance matrices [0.8,0.4;0.4,0.8], [0.8,0.4;0.4,0.8] and [0.8,-0.4;-0.4,0.8]. 
The number of points in three clusters are 2,100 and 2. 
As illustrated in \autoref{fig:extreme}, Fuzzy-ITISC successfully separates the three clusters
while the other three models regard the two minority clusters as a single entity. 
\autoref{tab:MBD} shows that the Max-BoundaryDist of Fuzzy-ITISC($T_2=0.1$) is much smaller 
than the other models. 
This finding reveals that Fuzzy-ITISC treats every observation as a valid observation, not outlier, 
and takes them into account. 
In other words, Fuzzy-ITISC is a \textit{fair} clustering algorithm 
which \textit{highlights} the boundary observations.
The degree to which Fuzzy-ITISC highlights the boundary points
depends on the temperature $T_2$, which will be discussed in \Cref{T2}.

\subsection{Importance Sampling Weight} \label{ISWeights}
\begin{figure*}
  \begin{center}
  \includegraphics[width=.9999\linewidth]{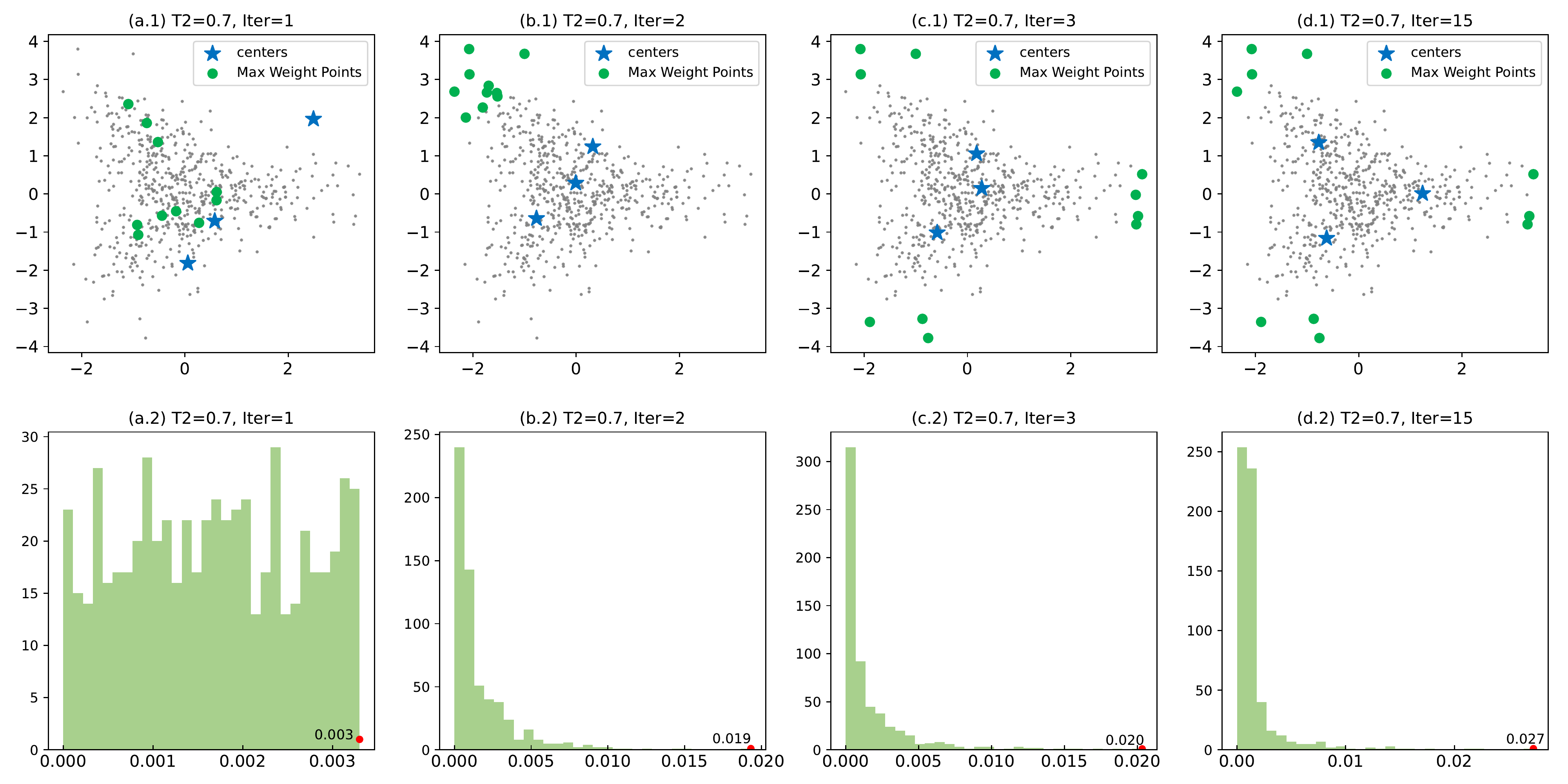}
  \end{center}
\caption{(1) The first row displays the top-10 weight points in Fuzzy-ITISC-AO($T_2=0.7$) at iteration 1,2,3,15.
(2) The second row displays the weight distribution at iteration 1,2,3,15.
The maximum weight is marked by a red point.
The model was trained on the default dataset with $T_2=0.7$ and converged at the iteration 15.}
\label{fig:ISWeights}
\end{figure*}
In this experiment, we investigate the properties of the importance sampling weight in Fuzzy-ITISC-AO.
We run the model on the default dataset with three clusters using Fuzzy-ITISC-AO at temperature $T_2=0.7$.
The $\epsilon$ in stopping criterion in \autoref{alg:ITISC-AO} is set to 1e-5
and the algorithm converged at the iteration 15.
The first row in \autoref{fig:ISWeights} displays the change of the top-10 maximum weights 
at iteration 1,2,3 and 15.
The figure indicates that the model gradually increases the bias in favor of boundary points. 
The second row in \autoref{fig:ISWeights} displays the weight distribution at iteration 1,2,3 and 15.
The maximum weights at the four iterations are 0.003, 0.019, 0.020 and 0.027,
which means the maximum weight gradually becomes larger. 
Furthermore, the weight distribution becomes more biased.
These trends indicate that the model concentrates on a small subset of the observations. 

\subsection{Effect of the Temperature $T_2$} \label{T2} 
\begin{figure}
  \begin{center}
  \includegraphics[width=.9999\linewidth]{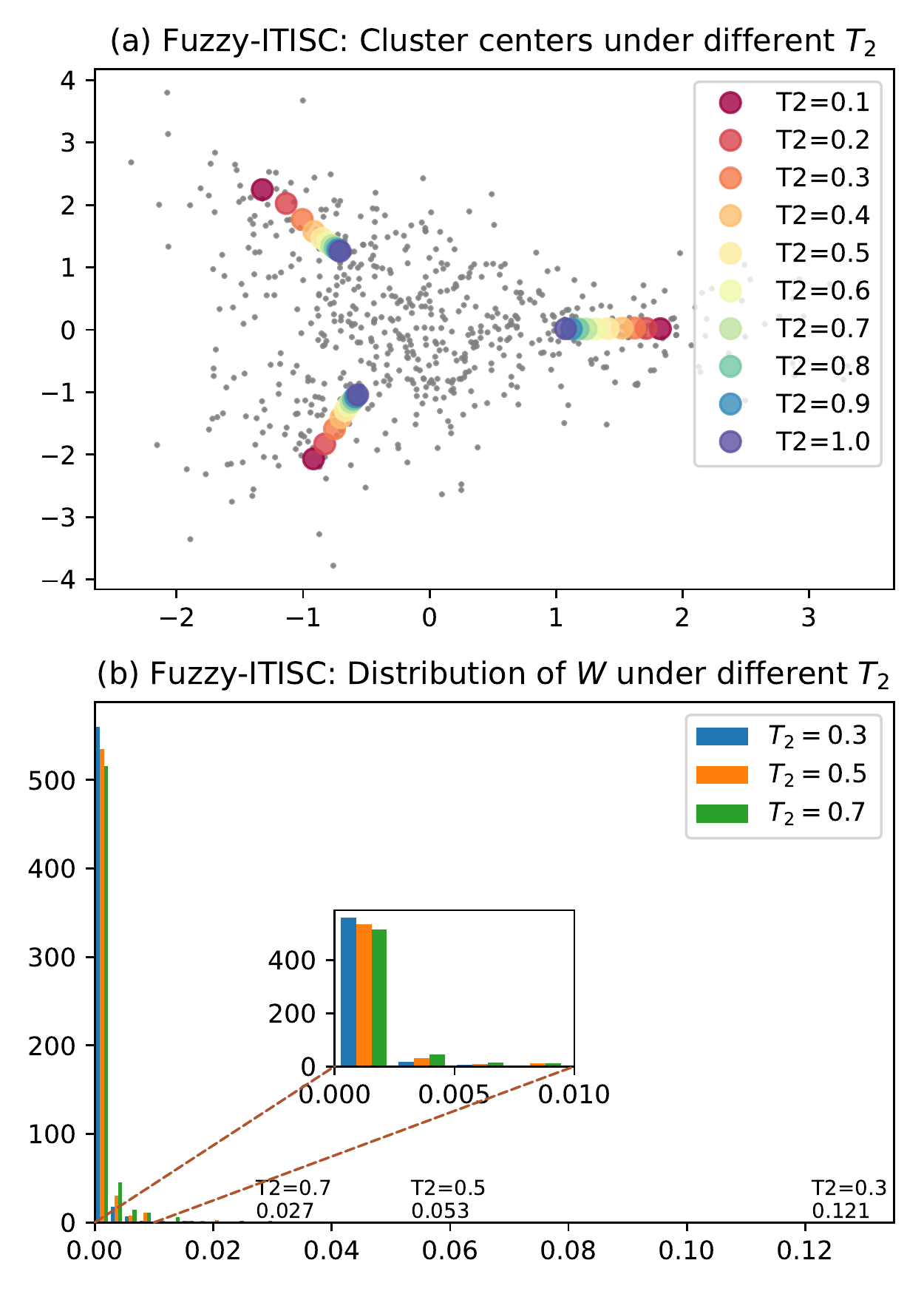}
  \end{center}
\caption{(a) Cluster centers of Fuzzy-ITISC under different $T_2$.
(b) Comparison of weight distributions among different $T_2$.
The annotation shows the maximum weight under different $T_2$.
The dashed lines zoom in the weight distribution between [0,0.01].}
\label{fig:diffT2}
\end{figure}

\begin{table}
  \caption{\label{tab:MBD-diffT2}
  Comparison of Max-BoundaryDist and 10-BoundaryDist under different $T_2$.
  $C=c$ means the Gaussian synthetic dataset with $c$ centers.}
  \centering
  \small
  \begin{tabular}{lcccccccc}
    \hline
      \multicolumn{1}{c|}{}  & \multicolumn{4}{c|}{Max-BoundaryDist} & \multicolumn{4}{c}{10-BoundaryDist} \\
      \multicolumn{1}{c|}{$T_2$}&C=2&C=3&C=4&\multicolumn{1}{c|}{C=6}&C=2&C=3&C=4&C=6 \\
      \hline
      \multicolumn{1}{c|}{2.0}&2.99&2.97&3.11&\multicolumn{1}{c|}{2.97}&23.35&26.24&24.81&27.57\\
      \multicolumn{1}{c|}{1.5}&2.94&2.96&3.06&\multicolumn{1}{c|}{2.92}&23.18&25.65&24.56&26.47\\
      \multicolumn{1}{c|}{1.0}&2.84&2.88&2.98&\multicolumn{1}{c|}{2.78}&22.68&24.48&23.95&25.48\\
      \multicolumn{1}{c|}{0.9}&2.81&2.86&2.94&\multicolumn{1}{c|}{2.74}&22.51&24.13&23.73&24.85\\
      \multicolumn{1}{c|}{0.8}&2.78&2.82&2.90&\multicolumn{1}{c|}{2.70}&22.31&23.72&23.44&24.08\\
      \multicolumn{1}{c|}{0.7}&2.74&2.77&2.85&\multicolumn{1}{c|}{2.65}&22.07&23.22&23.06&23.14\\
      \multicolumn{1}{c|}{0.6}&2.69&2.72&2.78&\multicolumn{1}{c|}{2.57}&21.75&22.60&22.55&21.92\\
      \multicolumn{1}{c|}{0.5}&2.62&2.63&2.69&\multicolumn{1}{c|}{2.43}&21.32&21.78&21.84&20.47\\
      \multicolumn{1}{c|}{0.4}&2.53&2.51&2.55&\multicolumn{1}{c|}{2.24}&20.70&20.63&20.78&18.77\\
      \multicolumn{1}{c|}{0.3}&2.39&2.29&2.31&\multicolumn{1}{c|}{1.89}&19.77&19.01&19.08&16.76\\
      \multicolumn{1}{c|}{0.2}&2.22&2.01&1.98&\multicolumn{1}{c|}{1.63}&18.51&16.97&16.67&14.57\\
      \multicolumn{1}{c|}{0.1}&2.08&1.72&1.66&\multicolumn{1}{c|}{1.37}&17.49&14.91&14.10&12.84\\
    \hline
  \end{tabular}
\end{table}

In this section, we examine the effect of the temperature $T_2$ on cluster centers and weight distributions. 
\autoref{fig:diffT2}(a) compares centers of Fuzzy-ITISC, FCM, k-means
and hierarchical clustering under different $T_2$.
The figure shows that as $T_2$ decreases from 1.0 to 0.1, 
the centers of Fuzzy-ITISC moving toward the boundary points.
Meanwhile, \autoref{tab:MBD-diffT2} compares the numeric results 
of Max-BoundaryDist and 10-BoundaryDist at different $T_2$
which takes values from $\{0.1,0.2,0.3,0.4,0.5,0.6,0.7,0.8,0.9,1.0,1.5,2.0\}$.
The table shows a decrease in both Max-BoundaryDist and 10-BoundaryDist as $T_2$ gets smaller,
indicating that the cluster centers become closer to the boundary points. 
In addition, \autoref{fig:diffT2}(b) displays how $T_2$ affects the weight distributions in Fuzzy-ITISC.
The figure shows that smaller $T_2$ leads to more sharply peaked weight distributions
while larger $T_2$ leads to broader weight distributions.
The maximum weights of three models under $T_2=0.7$, $T_2=0.5$, $T_2=0.3$ in Fuzzy-ITISC are
0.027, 0.053 and 0.121 respectively.
One possible explanation is that as $T_2 \rightarrow 0$, 
the constraint on $KL(w(x_i) \parallel \{\frac{1}{N}\})$ is small,
then $KL(w(x_i) \parallel \{\frac{1}{N}\})$ could be very large. 
Thus, smaller $T_2$ leads to more sharply peaked weight distributions.

\subsection{Experiment Results on Data Distribution Shifts} \label{dist-shift}
In the preceding section, we validate that the centers of Fuzzy-ITISC are closer to the boundary points 
compared with FCM, k-means and HC when $T_2$ is small. 
In this section, we simulate possible future distribution shifts by generating shifted Gaussian distributions 
and show that Fuzzy-ITISC performs better when the distribution shift is large. 
The distance between the original and the shifted Gaussian distributions is calculated by the KL divergence. 
The KL divergence between the two multivariate Gaussian distributions,
$\mathcal{N} (\mu_1, \Sigma_1)$ and $\mathcal{N} (\mu_2, \Sigma_2)$, 
is defined as follows\cite{duchi2007derivations}
\begin{align}
  \medmath{\text{KL-Dist} = \frac{1}{2} \bigl( log \frac{det \Sigma_2}{det \Sigma_1} - n + tr(\Sigma_2^{-1} \Sigma_1)
+ (\mu_2 - \mu_1)^{T} \Sigma_2^{-1} (\mu_2 - \mu_1)\bigr),}
\end{align} 
where $n$ is the number of dimensions of the data. 
This experiment explores two types of distribution shifts: 
mean translation and covariance matrix scaling.
For mean translation, a shifted distribution is generated from a new mean under the same covariance matrix. 
The new means are selected evenly on the circumference of the circle centered at the original mean $(a,b)$
with a radius of $S$. Here, we call $S$ the shifted mean distance
and larger $S$ implies larger distribution shifts.
The polar coordinate of the circle is defined as 
$x = S * cos(\phi) + a$ and $y = S * cos(\phi) + b$ where $\phi \in [0, 2\pi]$.
In this experiment, 13 equiangularly spaced points are selected,
therefore three Gaussian distributions in the default dataset lead to 13*13*13=2197 shifted Gaussian distributions in total.
For scale of the covariance matrix, the shifted distribution is generated from a scaled covariance matrix
by simply multiplying a scaling factor under the same mean.
The scaling factors for three clusters are $SF_1$, $SF_2$ and $SF_3$.
The 13 scaling factors are chosen from \{0.5,0.6,0.7,0.8,0.9,1.0,1.5,2,2.5,3,3.5,4,4.5\}.
The total KL divergence between the original and the shifted dataset is calculated by summing
three KL divergences together. 
\begin{figure}
  \begin{center}
  \includegraphics[width=.9999\linewidth]{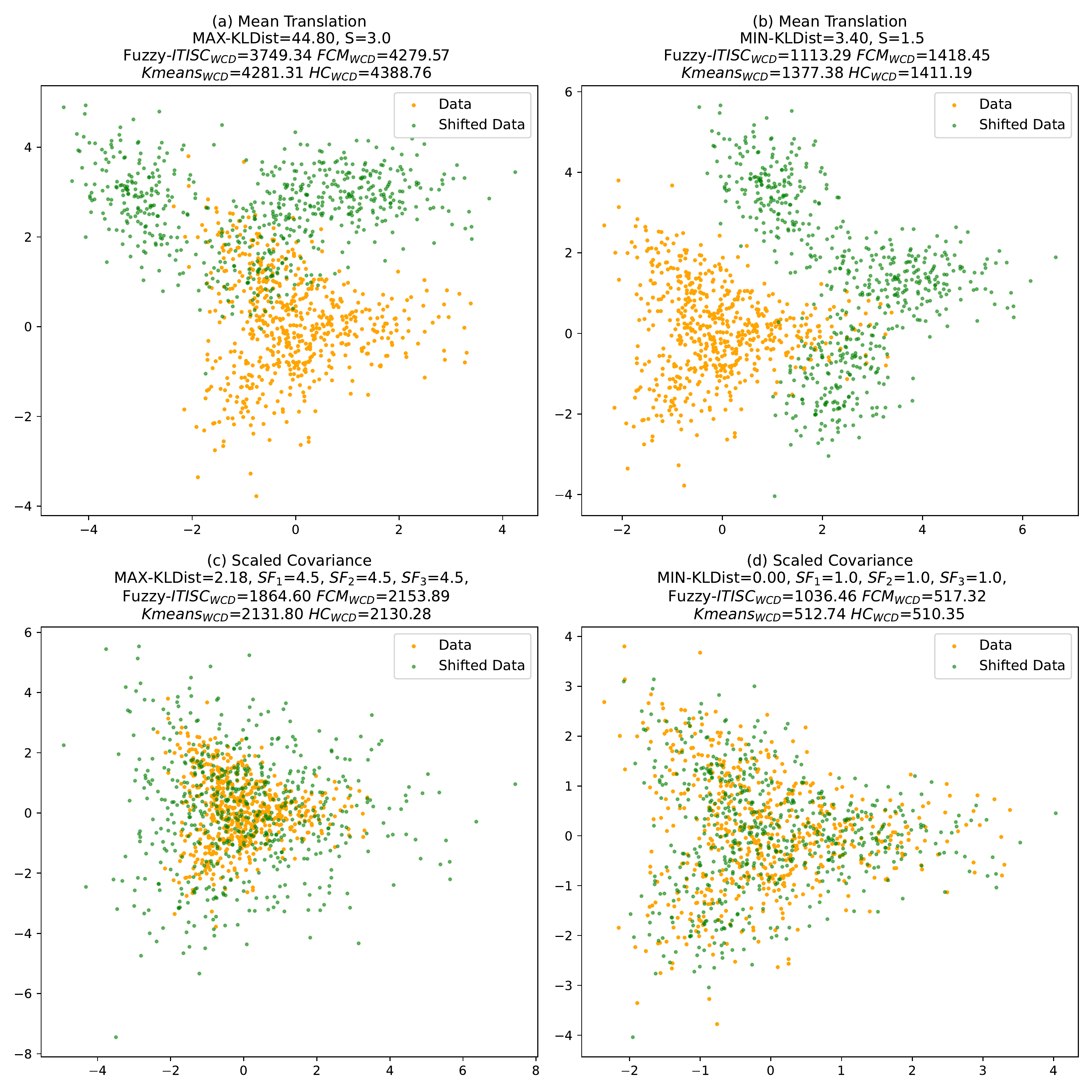}
  \end{center}
\caption{Original and shifted datasets under maximum and minimum KL divergence.
(a) and (b) show maximum and minimum distribution shifts under mean translation
where $S$ represents the shifted distance.
(c) and (d) show maximum and minimum distribution shifts under scaled covariance where
$SF_1$, $SF_2$ and $SF_3$ represent the scaling factors for three clusters. 
(d) shows the same distribution under a different random seed since all three covariance scaling factors are 1.0.
The subscript WCD represents WithinClusterDist.}
\label{fig:dist-shift}
\end{figure}
\begin{figure*}
  \begin{center}
  \includegraphics[width=.9999\linewidth]{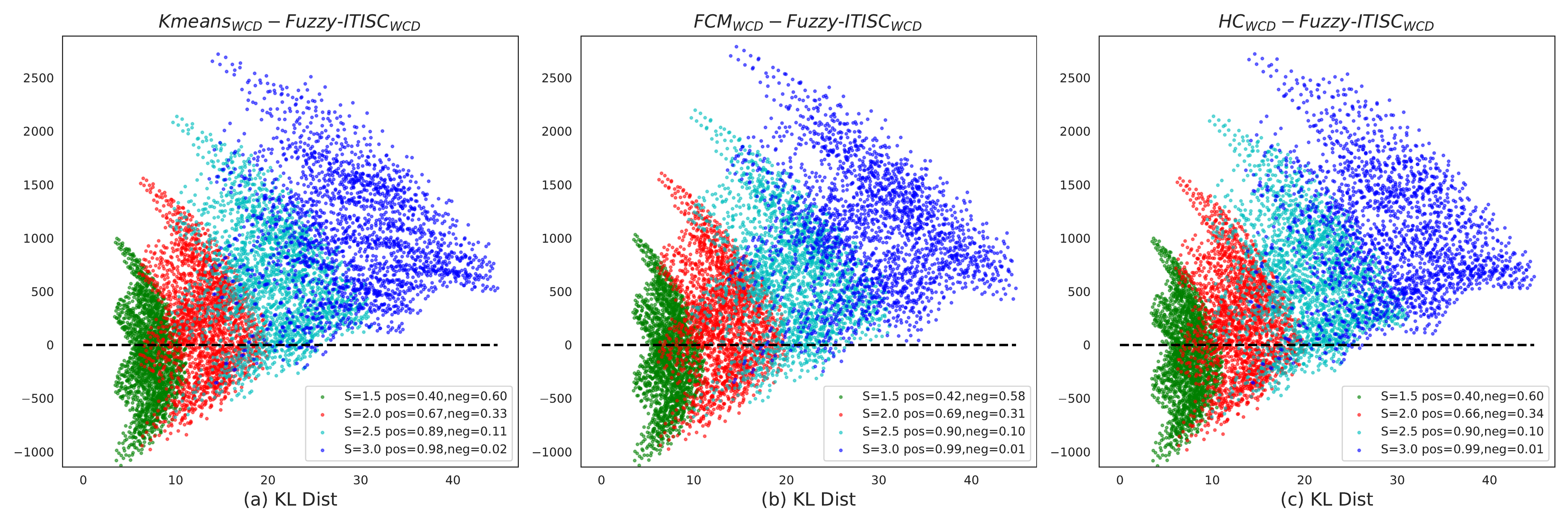}
  \end{center}
\caption{Comparison of WithinClusterDist(WCD) difference against KL divergence 
between the original and shifted distributions.  
X-axis displays the KL divergence between the original distribution
and the shifted distributions.
Y-axis displays the WithinClusterDist difference between k-means, FCM, HC and Fuzzy-ITISC. 
The black dotted horizontal line represents WithinClusterDist difference equals zero.
In the legend, $S$ represents shifted mean distance, 
\textit{pos} and \textit{neg} represent the ratio of positive and negative distance difference respectively.}
\label{fig:dist-shift-scatter}
\end{figure*} 

In this experiment, we first get four models (Fuzzy-ITISC($T_2=0.1$), FCM, k-means and HC) under the default dataset,  
then generating new datasets under the shifted distributions, 
next predicting on the shifted dataset and calculating the within-cluster-sum-of-distances, 
denoted as \textit{WithinClusterDist}.
The metric WithinClusterDist is used to measure 
\textit{how well a clustering model performs under a future distribution shift}, 
which is calculated by summing \textbf{all} distances within each cluster.
We calculated 2197 WithinClusterDist and show the maximum and the minimum ones
with respect to the KL divergence in \autoref{fig:dist-shift}.
WithinClusterDist of three clustering models are shown in the title of each subplot.
(a) and (b) illustrate the maximum and minimum distribution shifts under mean translations.
(c) and (d) illustrate the maximum and minimum distribution shifts under scaled covariances.
The figure shows that Fuzzy-ITISC performs better than FCM, k-means and HC 
when the distribution shift is large, as in (a), (b), (c);
meanwhile, it performs worse than FCM, k-means and HC 
when the distribution shift is small, as shown in (d).

Furthermore, \autoref{fig:dist-shift-scatter} compares the WithinClusterDist difference
between Fuzzy-ITISC and k-means (FCM,HC) against the KL divergence 
under different shift translation factor $S$.
In each subplot, we observe that larger values of $S$ leads to larger KL divergence,
indicating larger distribution shifts.
Points above the black dotted line indicate Fuzzy-ITISC performs better,
while points below the line indicate Fuzzy-ITISC performs worse. 
The ratios of positive and negative WithinClusterDist difference are displayed in the legend.
For values of S equal to \{1.5, 2.0, 2.5, 3.0\},
the ratios of Fuzzy-ITISC outperforming each algorithm were as follows: 
k-means -- \{0.40, 0.67, 0.89, 0.98\},
FCM -- \{0.42, 0.69, 0.90, 0.99\}, and
HC -- \{0.40, 0.66, 0.90, 0.99\}. 
Our results demonstrate that as the distribution shift becomes larger, 
Fuzzy-ITISC consistently outperforms other clustering algorithms, 
validating our assumption that Fuzzy-ITISC can \textit{do best in the worst case} 
where the level of ``worse" is measured by the KL divergence. 
In other words, Fuzzy-ITISC is effective in worst-case scenarios.

\section{Numerical Results on a Real-World Dataset} \label{load-forecasting}
In this section, we evaluate the performance of our proposed Fuzzy-ITISC algorithm
on a real-world load forecasting problem.
We first provide an overview of the method and then explain it in detail.  
Following \cite{fan2006short}, we use a two-stage approach. 
In the first stage, four clustering models(k-means, FCM, HC and Fuzzy-ITISC) 
are applied to separate the training days into $C$ clusters in an unsupervised manner. 
In the second stage, 
we use a Support Vector Regression\cite{smola2004tutorial} model
for each time stamp (a total of 96 time stamps)
to fit training data in each cluster in a supervised manner. 
For each test day, it is assigned to a cluster based on the trained clustering model, 
and for each time stamp, the corresponding regression model is used to predict the result. 
To evaluate the performance of the clustering model, 
we use the Mean Squared Error(MSE) on the test datasets,
where a smaller MSE implies a more reasonable separation of clusters.

\begin{figure}
  \begin{center}
  \includegraphics[width=.9999\linewidth]{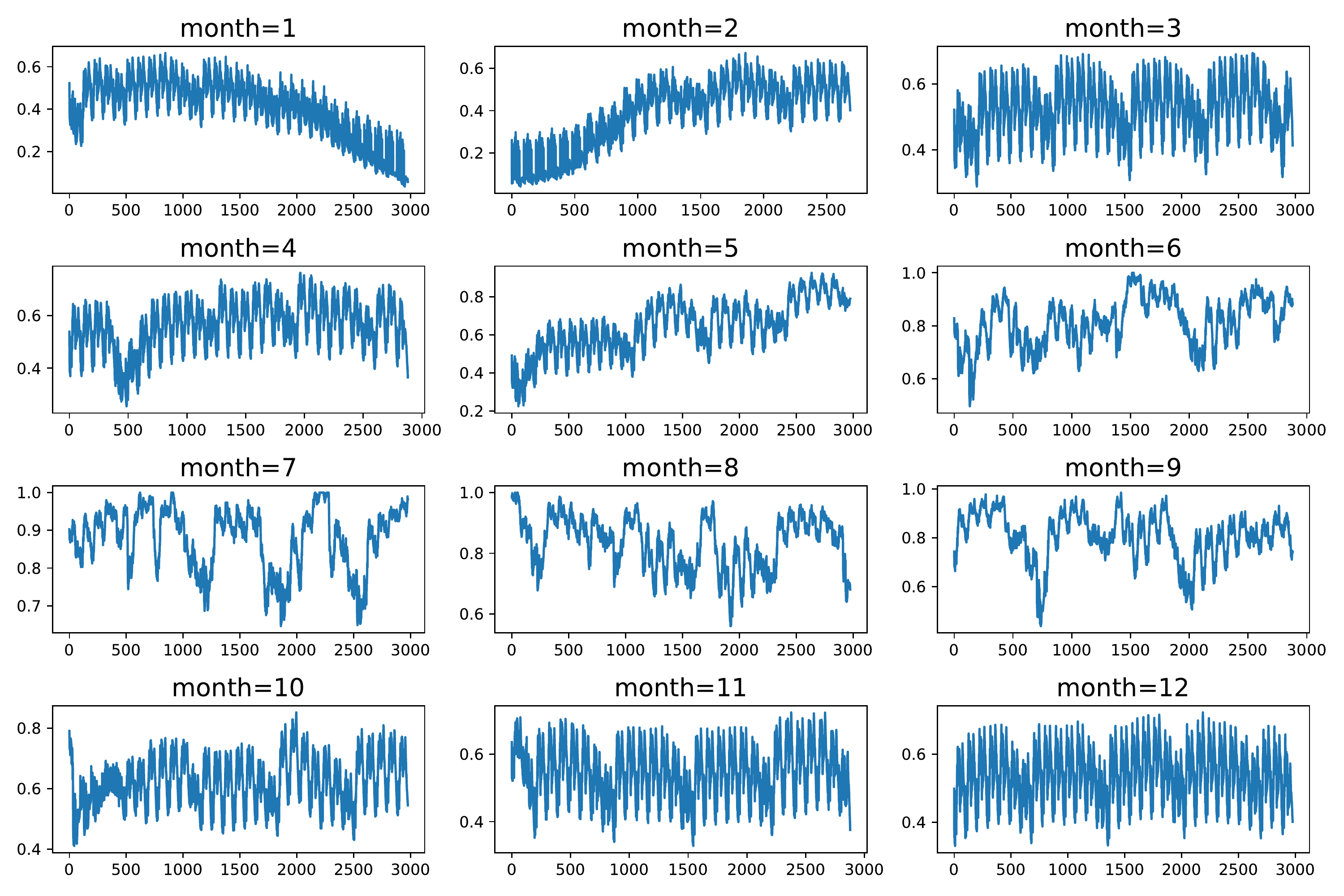}
  \end{center}
\caption{Normalized load in 2014 for each month on testing dataset. X-axis represents the time index.}
\label{fig:month-load}
\end{figure}

The load forecasting dataset \footnote{\url{http://shumo.neepu.edu.cn/index.php/Home/Zxdt/news/id/3.html}}
used in this section is from The Ninth Electrician Mathematical Contest in Modeling in China
and consists of two parts: historical loads and weather conditions.
Daily loads are recorded every 15 minutes, resulting in a total of 96 time stamps per day.
The weather dataset includes daily maximum, minimum, mean temperature, 
humid and rainfall. The time ranges from 20120101 to 20141231. 
We use the consecutive 24 months as the training dataset and the following one month as the test dataset. 
For instance, if the training dataset ranges from 20120201 to 20140131,
the corresponding test dataset is from 20140201 to 20140228.
There are 12 test datasets, from January to December.
Taking February as an example, the length of the training dataset is 731 and
the length of test dataset is 28. 
Therefore, the shapes of the training and test load data are [731,96] and [28,96] respectively. 
We normalize the training dataset for each time stamp using $\frac{x-x_{max}}{x_{max}-x_{min}}$,
and the test dataset is normalized using the statistics from the training dataset,
indicating that no future data is involved.
The normalized monthly load series are shown in 
\autoref{fig:month-load}.

Next, we describe the features used for the clustering and regression models. 
Inspired from \cite{fan2006short}, we use the following features for clustering:
previous day's maximum daily load,
last week's average maximum daily load and 
average of the previous two days' mean temperature.
Therefore, the shape of training feature for the clustering models is [731,3]. 
For regression models, we use historical loads 
from previous \{96, 100, 104, 192, 288, 384, 480, 576, 672\} time stamps,
resulting in a regression feature of length 9. 
We use Support Vector Regression(SVR) implemented by the sklearn\cite{scikit-learn} 
package as the regression model. 
In this example, we use $T_2=0.1$ as the default value in Fuzzy-ITISC.
For k-means, FCM and Fuzzy-ITISC, we use different random seeds 
for initialization and report the mean of four runs.
For hierarchical clustering, we use the ward, complete\cite{kalkstein1987evaluation}, average, 
and single linkage and report the mean of the four models. 
Finally, we explain the training routine in the following Procedure.
The shape of $x_{cls}$ is [trainDays,3] and the shape of $x_{reg}^{i}$ is [$x_i$, 96, 9].
$t_{label,j}^{i}$ is the true label for test data in cluster $i$ and time stamp $j$.
\begin{algorithm}
\textbf{Input}: trainData(abbr. $x$), testData(abbr. $t$)\\
\textbf{Parameter}: C \\
\textbf{Output}: $\text{MSE}_{\text{All}}$
\begin{algorithmic}[1]
  \Procedure{Load Forecasting Clustering}{} 
  \State Generate clustering train data, $x_{cls}$.  [trainDays,3]. 
  \State Generate clustering test data,  $t_{cls}$.  [testDays,3].
  \For{$i \in \{1,2,\dots,C\}$} 
      \State Generate regression train data $x_{reg}^{i}$. [$x_i$, 96, 9]. 
      \State Generate regression test  data $t_{reg}^{i}$. [$t_i$, 96, 9].
      \For{$j \in \{1,2,\dots,96\}$} 
          \State Learn a $\text{SVR}^{i}_{j}$ based on $x_{reg,j}^{i}$.
          \State Predict on $t_{reg,j}^{i}$ based on $\text{SVR}^{i}_{j}$.
          \State Calculate MSE($t_{reg,j}^{i}$,$t_{label,j}^{i}$).
          \State $\text{MSE}_{\text{All}}$ = $\text{MSE}_{\text{All}}$ + MSE($t_{reg,j}^{i}$,$t_{label,j}^{i}$).
      \EndFor
  \EndFor
\EndProcedure
\end{algorithmic}
\end{algorithm}

\begin{figure}
  \begin{center}
  \includegraphics[width=.9999\linewidth]{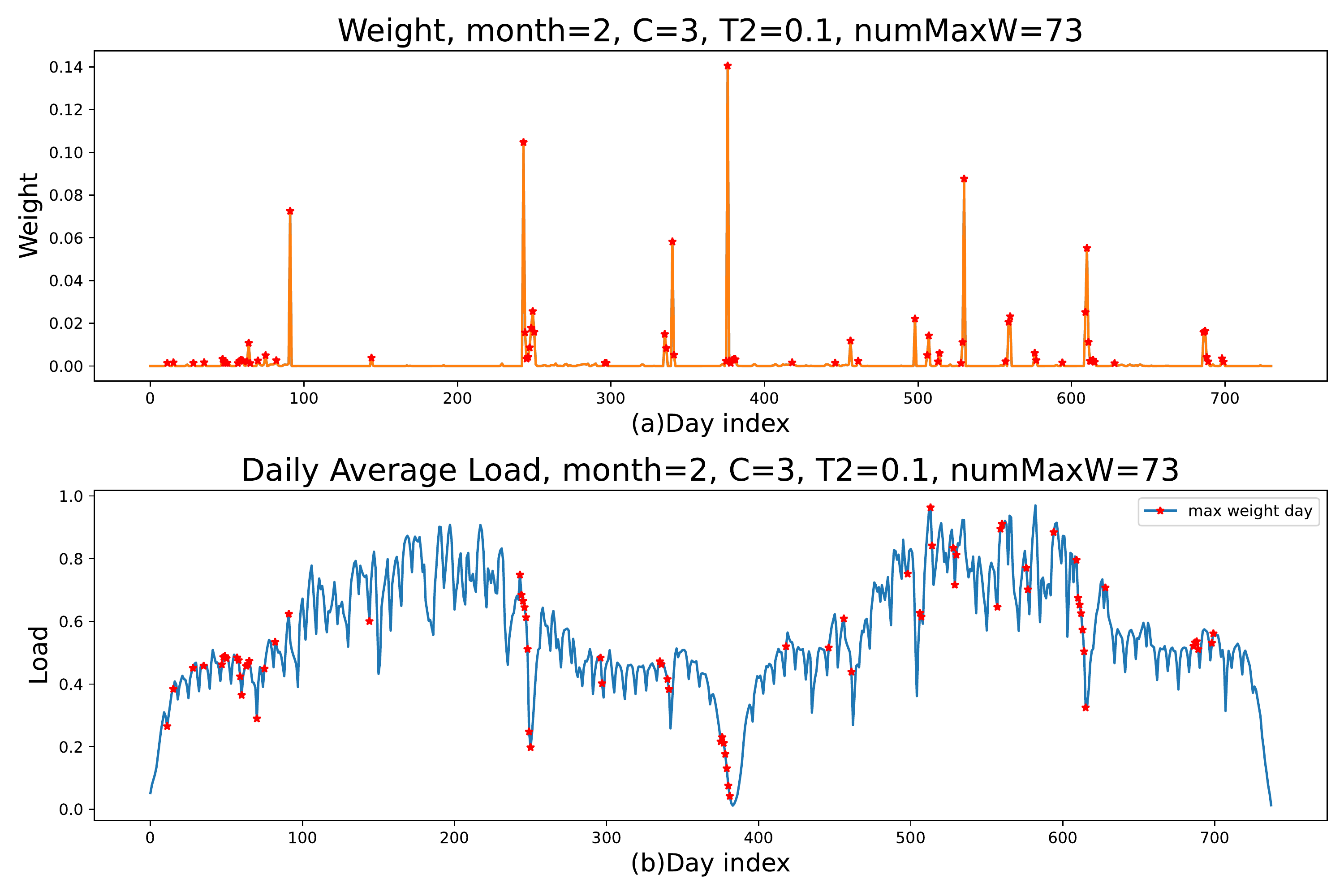}
  \end{center}
\caption{Fuzzy-ITISC($T_2=0.1$) weights and the corresponding average daily load for each training day.
The model was trained using three clusters in February.
The top 1\% weights and their corresponding daily loads are highlighted with red stars.
``numMaxW" represents the selected number of maximum weights.}
\label{fig:max-weight}
\end{figure}

Then, we present the experiment results on the load forecasting problem. 
First, \autoref{fig:max-weight} shows the importance sampling weight $W$ and 
the average daily loads of the training dataset. 
The model Fuzzy-ITISC($T_2=0.1$) is trained with $C=3$ clusters on February.
(a) shows Fuzzy-ITISC weights for each day and (b) shows each day's average daily load.
The red stars highlight the largest 1\% weights and their corresponding daily loads.
(b) indicates that daily loads with higher weights are around the valleys, which are data points with extreme values. 
This is because the clustering features partly rely on a day's previous daily loads.
Therefore, the data points around valleys and preceding the valleys are of higher weights. 
This observation indicates that Fuzzy-ITISC algorithm assigns higher weights to data points with extreme values. 

\begin{figure}
  \begin{center}
  \includegraphics[width=.9999\linewidth]{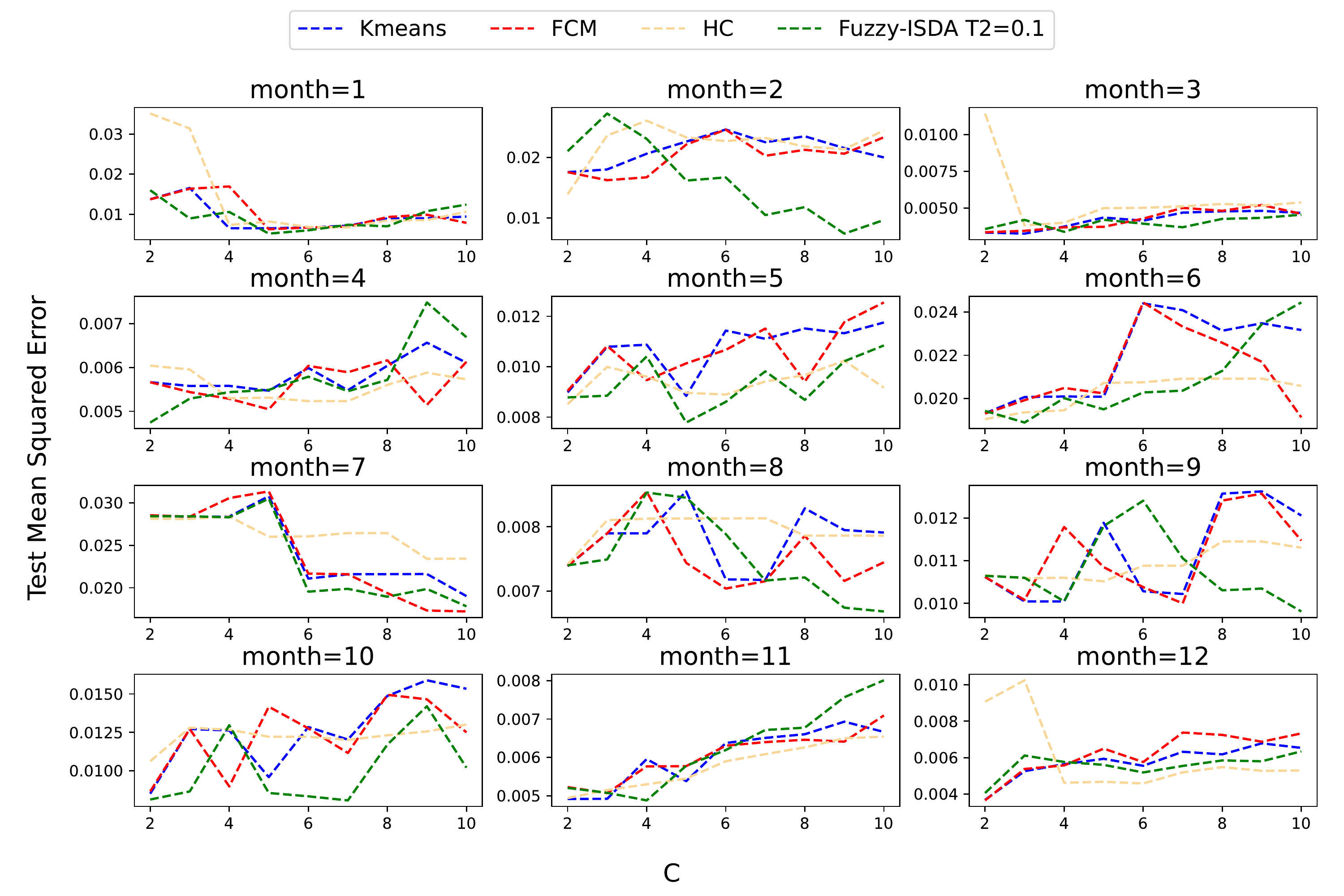}
  \end{center}
\caption{Comparison of test MSE for k-means, FCM, HC and Fuzzy-ITISC($T_2=0.1$) for each month.
X-axis displays the number of clusters $C$.
Y-axis displays the average test MSE of four runs.}
\label{fig:month-mse}
\end{figure}

Second, we compare the performance of k-means, FCM($m=2$) and Fuzzy-ITISC($T_2=0.1$) on test dataset.
For each clustering model, \autoref{tab:power-test-mse} reports the lowest test MSE 
among the models with varying numbers of clusters, ranging from $C=2$ to $C=10$. 
The results show that Fuzzy-ITISC($T_2=0.1$) outperforms k-means, FCM and HC 
in 10(1,2,4,5,6,8,9,10,11,12) out of 12 months.
\autoref{fig:month-mse} provides a detailed comparison 
of the test MSE for these models over 12 months and 9 clusters.
As the distribution of test dataset differs from the training dataset,
results in \autoref{fig:month-mse} and \autoref{tab:power-test-mse} 
demonstrate the effectiveness of Fuzzy-ITISC under future distribution shifts in most scenarios.

\begin{table}
  \caption{\label{tab:power-test-mse}
  Comparison of the best test MSE(shown in bold) for Fuzzy-ITISC ($T_2=0.1$), k-means, FCM and HC over 12 months.
  The best test MSE is the lowest test MSE among models with different number of clusters.}
  \centering
  \small
  \begin{tabular}{lcccc}
      \hline
        Month & K-means & FCM & HC & Fuzzy-ITISC \\
      \hline
        1 & 0.006485 & 0.006254 & 0.006724 & \textbf{0.005157} \\
        2 & 0.017573 & 0.016247 & 0.013928 & \textbf{0.007407} \\
        3 & \textbf{0.003275} & 0.003368 & 0.003831 & 0.003398 \\
        4 & 0.005470 & 0.005045 & 0.005232 & \textbf{0.004742} \\
        5 & 0.008842 & 0.009042 & 0.008512 & \textbf{0.007784} \\
        6 & 0.019318 & 0.019131 & 0.019049 & \textbf{0.018890}  \\ 
        7 & 0.019012 & \textbf{0.017205} & 0.023416 & 0.017812 \\ 
        8 & 0.007176 & 0.007042 & 0.007415 & \textbf{0.006686} \\ 
        9 & 0.010046 & 0.010002 & 0.010518 & \textbf{0.009811} \\ 
        10 & 0.008517 & 0.008667 & 0.010627 & \textbf{0.008081} \\ 
        11 & 0.004919 & 0.005071 & 0.004941 & \textbf{0.004880} \\ 
        12 & 0.003682 & 0.003654 & 0.004583 & \textbf{0.004059} \\
    \hline
\end{tabular}
\end{table}

\section{Conclusion} \label{conclusion}
In this paper, we propose a novel information theoretical importance sampling clustering method
to address the problem of distribution deviation between training and test data. 
The objective function of ITISC is derived from an information-theoretical viewpoint and 
\autoref{Fuzzy-ITISC-FCM} reveals that FCM is a special case of ITISC 
and the fuzzy exponent $m$ can be interpreted as the recalibration of temperature in thermodynamic system.
This observation provides a solid theoretical rationale for FCM.
Experiment results show that Fuzzy-ITISC outperforms k-means, FCM and hierarchical clustering
in worst-case scenarios on both synthetic and real-world datasets.
Furthermore, our proposed ITISC has several potential applications, 
such as designing deliver systems considering not only economic benefits but also accessibility to rural areas,
creating recommendation systems for users with few ratings and 
developing fair face recognition systems which takes care of the minority.
We plan to investigate these applications in our future work.

\section*{Disclosure statement}
No potential conflict of interest was reported by the authors.
\section*{Acknowledgments}
This work is supported by the National Natural Science Foundation of China under Grants 61976174.
Lizhen Ji is addtionally supported by the Nature Science Basis Research Program of Shaanxi (2021JQ-055),
the Ministry of Education of Humanities and Social Science Project of China (No.22XJCZH004), 
the Nature Science Basis Research Program of Shaanxi (No. 2023-JC-QN-0799)
and the Scientific Research Project of Shaanxi Provincial Department of Education (No.22JK0186).

\bibliographystyle{IEEEtran}  
\bibliography{ITISC}

\clearpage
\appendices 
\section{Synthetic Data set}
In this section, we present the synthetic data sets used in the main paper.
The data points within each cluster are normally distributed over a two-dimensional space.
There are 200 data points in each cluster by default. 
(a) shows a Gaussian dataset with two clusters.
Their respective means and covariance matrices are as follows 
\begin{flalign*}
\text{mean1}  &= [1.0, 0] \\
\text{mean2}  &= [-1.0,0] \\
\text{conv1}  &= [0.65,0.35;0.35,0.65] \\
\text{conv2}  &= [0.65,-0.35;-0.35,0.65] & 
\end{flalign*}
(b) shows a Gaussian dataset with three clusters.
Their respective means and covariance matrices are as follows 
\begin{flalign*}
\text{mean1} =& [1.0, 0] \\
\text{mean2} =& [-0.578,-1.0] \\
\text{mean3} =& [-0.578, 1.0] \\
\text{conv1} =& [1.0,0.0;0.0,0.3] \\
\text{conv2}  =& [0.475,0.303;0.303,0.825] \\
\text{conv3}  =& [0.475,-0.303;-0.303,0.825] &
\end{flalign*}
This dataset is called the \textit{default} dataset in the main paper.
(c) shows a Gaussian dataset with four clusters.
Their respective means and covariance matrices are as follows 
\begin{flalign*}
\text{mean1} =& [1.0, 1.0] \\
\text{mean2} =& [1.0,-1.0] \\
\text{mean3} =& [-1.0, -1.0] \\
\text{mean4} =& [-1.0,1.0] \\
\text{conv1} =& [0.55,0.45;0.45,0.55] \\
\text{conv2} =& [0.55,-0.45;-0.45,0.55] \\
\text{conv3} =& [0.55,0.45;0.45,0.55]   \\
\text{conv4} =& [0.55,-0.45;-0.45,0.55] &
\end{flalign*}
(d) shows a Gaussian dataset with six clusters.
Their respective means and covariance matrices are as follows 
\begin{flalign*}
\text{mean1}  =& [0.5, 0.867] \\
\text{mean2}  =& [-0.5,0.867] \\
\text{mean3}  =& [-1.0,0.0] \\
\text{mean4}  =& [-0.5,-0.867] \\
\text{mean5}  =& [0.5, -0.867] \\
\text{mean6}  =& [1.0, 0.0] \\
\text{conv1}  =& [0.475,0.303;0.303,0.825] \\
\text{conv2}  =& [0.475,-0.303;-0.303,0.825] \\
\text{conv3}  =& [1.0,0.0;0.0,0.3] \\
\text{conv4}  =& [0.475,0.303;0.303,0.825] \\
\text{conv5}  =& [0.475,-0.303;-0.303,0.825] \\
\text{conv6}  =& [1.0,0.0;0.0,0.3] &
\end{flalign*}

\section{Derivation of ITISC objective function} 
This appendix presents the derivations of the empirical estimation of expected
distortion, conditional entropy and KL-divergence using the idea of importance sampling.
Suppose $X$ is a discrete random variable with probability mass function $p(x)$, i.e., $p(x)=P(X=x)$. 
Suppose $q(x)$ is another discrete distribution such that $q(x)=0$ implies $f(x)p(x)=0$.
Then the expected distortion is
\begin{align}
  \begin{split}
     D & = \sum_x\sum_y p(x,y)d(x,y)  \\
       & = \sum_x p(x) \sum_y p(y|x)d(x,y) \\
       & = \frac{\sum_x \frac{p(x)}{q(x)}q(x) \sum_y p(y|x)d(x,y)}{\sum_x \frac{p(x)}{q(x)}q(x)} \\
       & = \sum_x \frac{\frac{p(x)}{q(x)}}{\sum_x \frac{p(x)}{q(x)}q(x)} \sum_y p(y|x)d(x,y) q(x),
  \end{split}
\end{align}
the conditional entropy is
\begin{align}
  \begin{split}
  & H(Y|X) = -\sum_x p(x)\sum_y p(y|x)log p(y|x) \\
  & = -\frac{\sum_x \frac{p(x)}{q(x)}q(x)\sum_y p(y|x)logp(y|x)}{\sum_x \frac{p(x)}{q(x)} q(x)} \\
  & = -\sum_x \frac{\frac{p(x)}{q(x)}}{\sum_x \frac{p(x)}{q(x)} q(x)} \sum_y p(y|x)logp(y|x) q(x),
  \end{split}
\end{align}
and the KL-divergence is
\begin{align}
  \begin{split} \label{eqn:KL-W}
  & KL(p(x) \parallel q(x)) =  \sum_x p(x) log \frac{p(x)}{q(x)} \\ 
  & = \sum_x \frac{\frac{p(x)}{q(x)}}{\sum_x \frac{p(x)}{q(x)} q(x)} 
      log \frac{\frac{p(x)}{q(x)}}{\sum_x \frac{p(x)}{q(x)} q(x)} q(x).
  \end{split}
\end{align}
Next, we derive the empirical estimation of the expected distortion, conditional entropy and KL-divergence.
Suppose $\{x_1,x_2,\cdots,x_N\}$ are $N$ i.i.d. samples drawn from $q(x)$.
The self-normalized importance sampling weight for $x_i$ is $w(x_i)$, which is 
\begin{equation}
w(x_i) = \frac{\frac{p(x_i)}{q(x_i)}}{\sum_{l=1}^{N} \frac{p(x_l)}{q(x_l)}}.
\end{equation}
$w(x_i)$ is denoted as $w_i$ for notation simplicity. 
Suppose the density for the discrete uniform distribution with $N$ points 
is denoted as $\{ \frac{1}{N} \}$ and the fuzzy membership $p(y_j|x_i)$ is denoted as $u_{ij}$.
Then we have the empirical estimation of the expected distortion
\begin{align}
  \label{eqn:ITISC-L-proof}
  \begin{split}
    \hat{D}   & = \frac{1}{N} \sum_{i=1}^{N} \frac{\frac{p(x_i)}{q(x_i)}}{\frac{1}{N} \sum_{l=1}^{N} \frac{p(x_l)}{q(x_l)}} 
            \sum_{j=1}^{C} p(y_i|x_i) d(x_i, y_j) \\
       &  = \sum_{i=1}^{N} w_i \sum_{j=1}^{C} u_{ij} d(x_i, y_j),
  \end{split}
\end{align}
the empirical estimation of the conditional entropy 
\begin{align}
  \label{eqn:ITISC-HYX-proof}
  \begin{split}
         & \hat{H}(Y|X) = -\frac{1}{N} \sum_{i=1}^{N} \frac{\frac{p(x_i)}{q(x_i)}}{\frac{1}{N} \sum_{l=1}^{N} \frac{p(x_l)}{q(x_l)}} 
              \sum_{j=1}^{C} p(y_j|x_i) log p(y_j|x_i)\\
         & = - \sum_{i=1}^{N} w_i \sum_{j=1}^{C} u_{ij} log u_{ij},
  \end{split}
\end{align}
and the empirical estimation of the KL-divergence
\begin{align}
  \label{eqn:ITISC-KLpq-proof}
  \begin{split}
  & \widehat{KL}(p(x) \parallel q(x)) = \frac{1}{N} \sum_{i=1}^{N} \frac{ \frac{p(x_i)}{q(x_i)} }{\frac{1}{N}\sum_{l=1}^{N} \frac{p(x_l)}{q(x_l)}} 
     log \frac{ \frac{p(x_i)}{q(x_i)} }{\frac{1}{N}\sum_{l=1}^{N} \frac{p(x_l)}{q(x_l)}} \\
  & = \sum_{i=1}^{N} \frac{ \frac{p(x_i)}{q(x_i)} }{\sum_{l=1}^{N} \frac{p(x_l)}{q(x_l)}}
      \left( log\big( \frac{ \frac{p(x_i)}{q(x_i)} }{\sum_{l=1}^{N} \frac{p(x_l)}{q(x_l)}} \big) - log \frac{1}{N} \right) \\
  & = \sum_{i=1}^{N} w_i (log(w_i) - log\frac{1}{N}) \\
  & = \sum_{i=1}^{N} w_i log  \frac{w_i}{\frac{1}{N}}  \\
  & = \sum_{i=1}^{N} w_i log w_i + log N \\
  & = KL(\{ w \} \parallel \{ \frac{1}{N} \}).
  \end{split}
\end{align}

\section{Derivation of Reformulation of ITISC}
In this appendix, we derive the necessary optimality condition for $U$ and $W$,
the reformulation $F_{ITISC}(Y,W)$ and $F_{ITISC}(Y)$
and the alternative updating rule for the cluster center $Y$ in the ITISC-AO algorithm.
First, we derive the necessary optimality condition for the fuzzy partition matrix $U$. 
Differentiating $\mathcal{L}_{ITISC}$ with respect to $u_{ij}$, we get
\begin{equation} 
\frac{\partial{\mathcal{L}_{ITISC}}}{\partial u_{ij}} = w_i \{ d(x_i,y_j) + T_1 (log u_{ij} + 1)\} - \lambda_i.\label{eq:L-d-uij}
\end{equation}
Setting the derivative \eqref{eq:L-d-uij} to zero, we get
\begin{equation}
u_{ij} = exp(\frac{\frac{\lambda_i}{w_i}-T_1}{T_1})exp(-\frac{ d(x_i, y_j)}{T_1}). \label{eq:uij-exp}
\end{equation}
Since $\sum_{j=1}^{C} u_{ij} = 1$, we have
\begin{equation} 
exp(\frac{\frac{\lambda_i}{w_i}-T_1}{T_1}) = \frac{1}{\sum_{j=1}^{C} exp(-\frac{d(x_i, y_j)}{T_1})}, \label{eq:uij}
\end{equation}
putting \eqref{eq:uij} back into \eqref{eq:uij-exp}, we get
\begin{equation}    
u_{ij} =  \frac{exp(-\frac{d(x_i, y_j)}{T_1})}{\sum_{j=1}^{C} exp(-\frac{d(x_i, y_j)}{T_1})}. \label{eq:appendix-uij}
\end{equation}
\eqref{eq:appendix-uij} is called the necessary optimality condition for $U$.
Second, we derive the reformulation for $U$.
Substituting $u_{ij}$ into the objective function $\mathcal{L}_{ITISC}$, 
we have
\begin{align}
& R_{ITISC}(Y,W)= \nonumber \\
& \sum_{i=1}^{N} w_i \{  \sum_{j=1}^{C} u_{ij} [ d(x_i, y_j) + T_1 log \frac{exp(-\frac{d(x_i, y_j)}{T_1})}{\sum_{j=1}^{C} exp(-\frac{d(x_i, y_j)}{T_1})} ] \} \nonumber \\
& \quad -T_2\sum_{i=1}^{N} w_i log w_i - \lambda (\sum_{i=1}^{N} w_i-1) \nonumber \\
& = \sum_{i=1}^{N} w_i \{  \sum_{j=1}^{C} u_{ij} [d(x_i, y_j) - d(x_i, y_j) - T_1 log \sum_{j=1}^{C} exp(-\frac{d(x_i, y_j)}{T_1}] \} \nonumber \\
& \quad -T_2\sum_{i=1}^{N} w_i log w_i - \lambda (\sum_{i=1}^{N} w_i-1)\nonumber \\
& = \sum_{i=1}^{N} w_i [-T_1 log \sum_{j=1}^{C} exp(-\frac{d(x_i, y_j)}{T_1})] \nonumber \\
& \quad -T_2\sum_{i=1}^{N} w_i log w_i - \lambda (\sum_{i=1}^{N} w_i-1). \label{eq:appendix-ITISC-Y-W}
\end{align}
\eqref{eq:appendix-ITISC-Y-W} is called the reformulation for $U$.
Third, we derive the necessary optimality condition for the importance sampling weight $W$.
Differentiating $R_{ITISC}(Y,W)$ with respect to $w_i$ and setting the derivative to zero, we get 
\begin{equation}
-T_1 log \sum_{j=1}^{C} exp(-\frac{d(x_i, y_j)}{T_1}) - T_2 (log w_i + 1) - \lambda = 0,
\end{equation}
therefore
\begin{align}
& w_i =  exp[-\frac{T_1}{T_2}  log \sum_{j=1}^{C} exp(-\frac{d(x_i, y_j)}{T_1}) - \frac{\lambda+T_2}{T_2}] \nonumber \\
& =  exp(log [\sum_{j=1}^{C} exp(-\frac{d(x_i, y_j)}{T_1})]^{-\frac{T_1}{T_2}}) exp(-\frac{\lambda+T_2}{T_2}) \nonumber \\
& =  [\sum_{j=1}^{C}  exp(-\frac{d(x_i, y_j)}{T_1})]^{-\frac{T_1}{T_2}} exp(-\frac{\lambda+T_2}{T_2}). \label{eq:wi-exp}
\end{align}
Since $\sum_{l=1}^{N} w_l = 1$, we have 
\begin{equation}
exp(-\frac{\lambda+T_2}{T_2}) = \frac{1}{\sum_{l=1}^{N}  [\sum_{j=1}^{C}exp(-\frac{d(x_l, y_j)}{T_1})]^{-\frac{T_1}{T_2}}}, \label{eq:wi}
\end{equation}
putting \eqref{eq:wi} back into \eqref{eq:wi-exp}, we have
\begin{equation}
w_i = \frac{[\sum_{j=1}^{C}  exp(-\frac{d(x_i, y_j)}{T_1})]^{-\frac{T_1}{T_2}}}
{\sum_{l=1}^{N} [\sum_{j=1}^{C}  exp(-\frac{d(x_l, y_j)}{T_1})]^{-\frac{T_1}{T_2}}}.\label{eq:appendix-wi}
\end{equation}
\eqref{eq:appendix-wi} is called the necessary optimality condition for $W$.
Fourth, we derive the reformulation for $U$ and $W$.
Here we denote $\sum_{j=1}^{C} exp(-\frac{d(x_i, y_j)}{T_1})$ as $A_i$ for notation simplicity.
Substituting \eqref{eq:appendix-wi} into \eqref{eq:appendix-ITISC-Y-W}, we have
\begin{flalign}
  & R_{ITISC}(Y) =  -T_1 \sum_{i=1}^{N} \frac{A_i^{-\frac{T_1}{T_2}}}{\sum_{l=1}^{N} A_l^{-\frac{T_1}{T_2}}}log (A_i) \nonumber \\
  & -T_2 \sum_{i=1}^{N} \frac{A_i^{-\frac{T_1}{T_2}}}{\sum_{l=1}^{N} A_l^{-\frac{T_1}{T_2}}} log \frac{A_i^{-\frac{T_1}{T_2}}}{\sum_{l=1}^{N} A_l^{-\frac{T_1}{T_2}}} \nonumber \\
  = & -T_1 \sum_{i=1}^{N} \frac{A_i^{-\frac{T_1}{T_2}}}{\sum_{l=1}^{N} A_l^{-\frac{T_1}{T_2}}}log (A_i) \nonumber \\
    & -T_2 \sum_{i=1}^{N} \frac{A_i^{-\frac{T_1}{T_2}}}{\sum_{l=1}^{N} A_l^{-\frac{T_1}{T_2}}} log (A_i^{-\frac{T_1}{T_2}}) \nonumber \\
    & + T_2\sum_{i=1}^{N}\frac{A_i^{-\frac{T_1}{T_2}}}{\sum_{l=1}^{N} A_l^{-\frac{T_1}{T_2}}}  log(\sum_{l=1}^{N} A_l ^{-\frac{T_1}{T_2}})\nonumber \\
  = & T_2 log(\sum_{l=1}^{N} A_l ^{-\frac{T_1}{T_2}}) \nonumber \\
  = &  T_2 log(\sum_{l=1}^{N} [\sum_{j=1}^{C} exp(-\frac{d(x_i, y_j)}{T_1})]^{-\frac{T_1}{T_2}}). \label{eq:appendix-ITISC-Y}
\end{flalign}
\eqref{eq:appendix-ITISC-Y} is called the reformulation for $U$ and $W$.
Finally, we derive the update rule for the cluster center $Y$ in ITISC-AO algorithm. 
\begin{align}
  & \medmath{\frac{\partial}{\partial y_k} R_{ITISC}(Y)} \nonumber \\
  & = \medmath{\frac{\partial}{\partial y_k} T_2 log(\sum_{i=1}^{N} [\sum_{j=1}^{C} exp(-\frac{d(x_i, y_j)}{T_1})]^{-\frac{T_1}{T_2}})} \nonumber \\ 
  & = \medmath{\frac{\sum_{i=1}^{N}[\sum_{j=1}^{C} exp(-\frac{d(x_i, y_j)}{T_1})]^{-\frac{T_1}{T_2}-1}
      exp(-\frac{d(x_i, y_k)}{T_1}) \frac{\partial}{\partial y_k}d(x_i, y_k)}
      {\sum_{i=1}^{N}[\sum_{j=1}^{C} exp(-\frac{d(x_i, y_j)}{T_1})]^{-\frac{T_1}{T_2}}}} \nonumber \\
  & =  \medmath{\sum_{i=1}^{N} \biggl[\frac{[\sum_{j=1}^{C} exp(-\frac{d(x_i, y_j)}{T_1})]^{-\frac{T_1}{T_2}}}
       {\sum_{i=1}^{N}[\sum_{j=1}^{C} exp(-\frac{d(x_i, y_j)}{T_1})]^{-\frac{T_1}{T_2}}}
       \frac{exp(-\frac{d(x_i, y_k)}{T_1})}{\sum_{j=1}^{C} exp(-\frac{d(x_i, y_j)}{T_1})}} \nonumber \\
  & \medmath{ \quad \quad \quad \frac{\partial d(x_i, y_k)}{\partial y_k} \biggl]} \nonumber \\
  & = \medmath{\sum_{i=1}^{N} w_i u_{ik} \frac{\partial d(x_i, y_k)}{\partial y_k}}. 
\end{align}
Setting the derivative $\frac{\partial}{\partial y_k} R_{ITISC}(Y)$ to zero,
we get the necessary optimality condition for $y_k$, which is  
\begin{equation}
\sum_{i=1}^{N} w_i u_{ik} \frac{\partial d(x_i, y_k)}{\partial y_k} = 0. 
\end{equation}
For Euclidean distance, the update rule for $y_k$ is
\begin{equation}
  y_k = \frac{\sum_{i=1}^{N} w_i u_{ik} x_i}{\sum_{i=1}^{N} w_i u_{ik}}. 
\end{equation}

\section{Derivation of Fuzzy-ITISC-AO update rule}
In this appendix, we derive the update rule for cluster center $Y$ in Fuzzy-ITISC-AO algorithm. 
Setting $\frac{\partial}{\partial y_k} F_{ITISC}(Y)$ to zero,
we can get the necessary optimality condition for $y_k$.
\begin{align} 
  & \medmath{\frac{\partial}{\partial y_k} R^{fuzzy}_{ITISC}(Y)} \nonumber \\
  &  = \medmath{\frac{\partial}{\partial y_k} \medmath{T_2 log (\sum_{i=1}^{N}(\sum_{j=1}^{C} d(x_i, y_j)^{-\frac{1}{T_1}})^{-\frac{T_1}{T_2}})}} \nonumber\\
  &  = \medmath{T_2 \sum\limits_{i=1}^{N} \frac{ (-\frac{T_1}{T_2}) 
  (\sum\limits_{j=1}^{C} d(x_i, y_j)^{-\frac{1}{T_1}})^{-\frac{T_1}{T_2}-1} (-\frac{1}{T_1})d(x_i, y_k)^{-\frac{1}{T_1}-1}
  \frac{\partial}{\partial y_k} d(x_i,y_k)}
  {\sum\limits_{i=1}^{N} (\sum\limits_{j=1}^{C} d(x_i, y_j)^{-\frac{1}{T_1}})^{-\frac{T_1}{T_2}}}} \nonumber \\
  %
  &  = \medmath{\sum\limits_{i=1}^{N} \frac{ 
  (\sum\limits_{j=1}^{C} d(x_i, y_j)^{-\frac{1}{T_1}})^{-\frac{T_1}{T_2}-1} d(x_i, y_k)^{-\frac{1}{T_1}-1}
  \frac{\partial}{\partial y_k} d(x_i,y_k)}
  {\sum\limits_{i=1}^{N} (\sum\limits_{j=1}^{C} d(x_i, y_j)^{-\frac{1}{T_1}})^{-\frac{T_1}{T_2}}}} \nonumber \\
  &  = \medmath{\sum_{i=1}^{N} \frac{(\sum_{j=1}^{C} d(x_i, y_j)^{-\frac{1}{T_1}})^{-\frac{T_1}{T_2}}} 
       {\sum\limits_{i=1}^{N} (\sum\limits_{j=1}^{C} d(x_i, y_j)^{-\frac{1}{T_1}})^{-\frac{T_1}{T_2}}}
       \frac{d(x_i, y_k)^{-\frac{1}{T_1}-1}}
       {(\sum_{j=1}^{C} d(x_i, y_j)^{-\frac{1}{T_1}})}
       \frac{\partial}{\partial y_k}d(x_i,y_k)} \nonumber \\
  &  = \medmath{\sum_{i=1}^{N} \frac{(\sum_{j=1}^{C} d(x_i, y_j)^{-\frac{1}{T_1}})^{-\frac{T_1}{T_2}+T1}} 
  {\sum\limits_{i=1}^{N} (\sum\limits_{j=1}^{C} d(x_i, y_j)^{-\frac{1}{T_1}})^{-\frac{T_1}{T_2}}}
  \frac{(d(x_i, y_k)^{-\frac{1}{T_1}})^{1+T_1}}
  {(\sum_{j=1}^{C} d(x_i, y_j)^{-\frac{1}{T_1}})^{1+T_1}}} \nonumber \\
  & \quad \quad \quad \medmath{ \frac{\partial}{\partial y_k}d(x_i,y_k)} \nonumber \\ 
  &  = \medmath{\sum_{i=1}^{N} \frac{(\sum_{j=1}^{C} d(x_i, y_j)^{-\frac{1}{T_1}})^{-\frac{T_1}{T_2}+\frac{T_1}{T_2}T_2}} 
  {\sum\limits_{i=1}^{N} (\sum\limits_{j=1}^{C} d(x_i, y_j)^{-\frac{1}{T_1}})^{-\frac{T_1}{T_2}}}
  \frac{(d(x_i, y_k)^{-\frac{1}{T_1}})^{1+T_1}}
  {(\sum_{j=1}^{C} d(x_i, y_j)^{-\frac{1}{T_1}})^{1+T_1}}} \nonumber \\
  & \quad \quad \quad \medmath{\frac{\partial}{\partial y_k}d(x_i,y_k)} \nonumber \\
  &  = \medmath{\sum_{i=1}^{N} \frac{(\sum_{j=1}^{C} d(x_i, y_j)^{-\frac{1}{T_1}})^{-\frac{T_1}{T_2}(1-T_2)}} 
  {\sum\limits_{i=1}^{N} (\sum\limits_{j=1}^{C} d(x_i, y_j)^{-\frac{1}{T_1}})^{-\frac{T_1}{T_2}}}
  \frac{(d(x_i, y_k)^{-\frac{1}{T_1}})^{1+T_1}}
  {(\sum_{j=1}^{C} d(x_i, y_j)^{-\frac{1}{T_1}})^{1+T_1}}} \nonumber \\
  & \quad \quad \quad \medmath{ \frac{\partial}{\partial y_k}d(x_i,y_k)} \nonumber \\
  &  = \medmath{ \sum_{i=1}^{N} \frac{1}{(\sum\limits_{i=1}^{N} (\sum\limits_{j=1}^{C} d(x_i, y_j)^{-\frac{1}{T_1}})^{-\frac{T_1}{T_2}})^{T_2}}
  \frac{(( \sum_{j=1}^{C} d(x_i, y_j)^{-\frac{1}{T_1}})^{\frac{T_1}{T_2}})^{(1-T_2)}} 
  {(\sum\limits_{i=1}^{N} (\sum\limits_{j=1}^{C} d(x_i, y_j)^{-\frac{1}{T_1}})^{-\frac{T_1}{T_2}})^{(1-T_2)}}} \nonumber \\
  & \quad \quad \quad \medmath{ \frac{(d(x_i, y_k)^{-\frac{1}{T_1}})^{1+T_1}}
  {(\sum_{j=1}^{C} d(x_i, y_j)^{-\frac{1}{T_1}})^{1+T_1}}
  \frac{\partial}{\partial y_k}d(x_i,y_k)} \nonumber \\
  & =  \medmath{\sum_{i=1}^{N} 
        \frac{w_i^{(1-T_2)} u_{ik}^{(1+T_1)}\frac{\partial}{\partial y_k}d(x_i,y_k)}
        {(\sum\limits_{i=1}^{N} (\sum\limits_{j=1}^{C} d(x_i, y_j)^{-\frac{1}{T_1}})^{-\frac{T_1}{T_2}})^{T_2}}} \label{eq:appendix-Fuzzy-ITISC-necc}
\end{align}
In \eqref{eq:appendix-Fuzzy-ITISC-necc}, the denominator $(\sum\limits_{i=1}^{N} (\sum\limits_{j=1}^{C} d(x_i, y_j)^{-\frac{1}{T_1}})^{-\frac{T_1}{T_2}})^{T_2}$ is the 
summation over $i$ and $j$, therefore it is a constant. 
Then, the optimality condition for $y_k$ is as follows
\begin{equation}
\sum_{i=1}^{N} w_i^{(1-T_2)} u_{ik}^{(1+T_1)} \frac{\partial}{\partial y_k}d(x_i,y_k) = 0.
\end{equation}
For Euclidean distance, the update rule for center $y_k$ is
\begin{equation}
  y_k = \frac{\sum_{i=1}^{N} w_i^{1-T_2} u_{ik}^{1+T_1} x_i}{\sum_{i=1}^{N} w_i^{1-T_2} u_{ik}^{1+T_1}}.
\end{equation}
\end{document}